\documentclass[ijoc,sgblindrev]{informs4} % current default for manuscript submission
\usepackage{txfonts}
\usepackage{comment}

\DeclareMathVersion{bold}
\usepackage{algorithm}  
\usepackage{algpseudocode}
\usepackage{caption}
\usepackage{color}
\usepackage{subfigure}
\usepackage{appendix}
\usepackage{amsmath}
\usepackage{bm}
\usepackage{multirow} 

\OneAndAHalfSpacedXI
 \usepackage{booktabs}

\usepackage{natbib}
 \bibpunct[, ]{(}{)}{,}{a}{}{,}%
 \def\bibfont{\small}%

\usepackage[bookmarks=false]{hyperref}

\hypersetup{
   colorlinks=true,
   linkcolor=[RGB]{17, 14, 178}, %[RGB]{17, 14, 178}, %[RGB]{17, 14, 178},
   citecolor= [RGB]{17, 14, 178}, %[RGB]{17, 14, 178}, %,[RGB]{108, 108, 108}, 
   filecolor= [RGB]{17, 14, 178}, %[RGB]{17, 14, 178}, %[RGB]{17, 14, 178},      
   urlcolor= [RGB]{17, 14, 178}, %[RGB]{17, 14, 178},
	pdfpagemode=FullScreen,
   }

\newcommand{\oE}{\mathbb{E}}
\newcommand{\oV}{\mathbb{V}}
\newcommand{\mR}{\mathbb{R}}

\TheoremsNumberedThrough     % Preferred (Theorem 1, Lemma 1, Theorem 2)
\ECRepeatTheorems
\EquationsNumberedThrough    % Default: (1), (2), ...
\begin{document}
%%%%%%%%%%%%%%%%

%\setcounter{page}{1}
%\DOI{10.1287/xxxx.0000.0000}%

\RUNAUTHOR{Lin, Luo and Hong}

% \RUNTITLE{Predictive Maintenance in Manufacturing}
\RUNTITLE{Contextual Strongly Convex Simulation Optimization}

% \TITLE{Optimal Resource Allocation in Humanitarian Logistics: A Stochastic Programming Approach}
\TITLE{Contextual Strongly Convex Simulation Optimization: Optimize then Predict with Inexact Solutions
}

\ARTICLEAUTHORS{%
\AUTHOR{Nifei Lin\textsuperscript{a},
Heng Luo\textsuperscript{b},
L. Jeff Hong\textsuperscript{c}}
\AFF{\textsuperscript{a} School of Management, Fudan University, Shanghai 200433, China; 
\textsuperscript{b} School of Data Science, Fudan University, Shanghai 200433, China;\\
\textsuperscript{c} Department of Industrial and Systems Engineering, University of Minnesota, Minneapolis, MN 55455, USA\\
} 
\AFF{\textbf{Contact:} \EMAIL{nflin19@fudan.edu.cn} (NL); \EMAIL{hluo22@m.fudan.edu.cn} (HL); \EMAIL{lhong@umn.edu} (LJH)}
\vspace{-10pt}
} % end of the block

\ABSTRACT{
In this work, we study contextual strongly convex simulation optimization and adopt an ``optimize then predict'' (OTP) approach for real-time decision making. In the offline stage, simulation optimization is conducted across a set of covariates to approximate the optimal-solution function; in the online stage, decisions are obtained by evaluating this approximation at the observed covariate.  The central theoretical challenge is to understand how the inexactness of solutions generated by simulation-optimization algorithms affects the optimality gap, which is overlooked in existing studies. To address this, we develop a unified analysis framework that explicitly accounts for both solution bias and variance.  Using Polyak–Ruppert averaging SGD as an illustrative simulation-optimization algorithm, we analyze the optimality gap of OTP under four representative smoothing techniques: $k$ nearest neighbor, kernel smoothing, linear regression, and kernel ridge regression.  We establish convergence rates, derive the optimal allocation of the computational budget $\Gamma$ between the number of design covariates and the per-covariate simulation effort, and demonstrate the convergence rate can approximately achieve $\Gamma^{-1}$ under appropriate smoothing technique and sample-allocation rule.  Finally, through a numerical study, we validate the theoretical findings and demonstrate the effectiveness and practical value of the proposed approach.
}%

% We implement optimization algorithms on designed covariates and approximate the optimal-solution function with respect to covariate by smoothing techniques in the offline stage. Subsequently, the online optimization problem is transformed into an estimation problem. 

% \FUNDING{This research was supported by [grant number, funding agency].}

%Supplemental Material:
%Data Ethics & Reproducibility Note:

%\KEYWORDS{Stochastic programming, Decision support,Uncertainty, Disaster response, Optimization}

%\HISTORY{Received: Month DD, YYYY; Accepted: Month DD, YYYY; Published Online: Month DD, YYYY}

\maketitle
%%%%%%%%%%%%%%%%%%%%%%%%%%%%%%%%%%%%%%%%%%%%%%%%%%%%%%%%%%%%%%%%%%%%%%

% Text of your paper here
%\vspace{-15pt}
\section{Introduction}\label{sec:intro}

In many real-world applications, decision-makers face the challenge of managing complex systems by selecting appropriate values for controllable parameters to optimize system performance. Due to the presence of stochastic dynamics, system performance often lacks closed-form analytical expressions and therefore requires simulation models for estimation. To better understand system behavior under changing environment, simulation models often incorporate covariates, such as environmental factors, that influence system outputs. 
The optimization problem based on the outputs of such simulation models is known as contextual simulation optimization, and it can be formulated as:
\begin{equation}\label{pr:CSCSO}
\min_{\theta\in\Theta} f\left(\theta; x\right) :=\oE[F\left(\theta; x\right)],    
\end{equation}
where $\theta\in\Theta\subset \mathbb{R}^{q}$ denotes the decision variables, \( x \in \mathcal{X}\subset \mathbb{R}^{d}\) represents the covariate vector, $F\left(\theta; x\right)$  is the stochastic system performance observed through simulation, and the expectation is taken with respect to the randomness in the simulation process. In this paper, we focus on a particular class of problems: contextual
strongly convex simulation optimization (CSCSO) problems, where $f\left(\theta; x\right)$ is strongly convex in $\theta$ for any given $x$, and $\Theta\times\mathcal{X}$ is a convex and compact set in $\mathbb{R}^{q+d}$. Such problems frequently arise in areas such as operations management and machine learning \citep{chaudhuri2011differentially,lyu2024closing}. 

In practice, Problem~(\ref{pr:CSCSO}) is often solved after the covariates are observed, effectively reducing it to a traditional simulation-optimization (SO) problem. This process may need to be repeated whenever the covariates change significantly. A key limitation of this online approach is its inefficiency in real-time decision-making. Since simulations are typically computationally expensive, optimizing the objective function after covariate observation can cause substantial delays, rendering the approach impractical in time-sensitive environments. This motivates the development of more efficient approaches that incorporate contextual information while reducing the burden of online simulation.

To address the challenge of real-time simulation-based decision-making, \cite{hong2019offline} proposed the offline-simulation–online-application (OSOA) framework. This framework advocates shifting computationally expensive simulation experiments from the online stage to the offline stage, where ample time and computational resources are available. The learned performance function is then used to support decision-making in the online stage. The OSOA framework can also be adapted to contextual SO, where extensive offline simulation experiments are conducted to learn either the objective function $f\left(\theta; x\right)$ or the optimal-solution function $\theta_*\!\left(x\right)$. In both cases, once the covariates are observed at the online stage, no additional simulation is required, enabling fast, real-time decisions. These two approaches represent the primary paradigms for solving contextual SO problems, commonly referred to as ``predict-then-optimize'' (PTO) and ``optimize-then-predic'' (OTP).

Existing research on contextual SO largely aligns with the two paradigms introduced above. The first stream follows the PTO approach. In this setting, simulation experiments are conducted at various covariate realizations during the offline stage to learn an approximation of the objective function. Upon observing a new covariate in the online stage, one first ``predicts" the objective value or function and then ``optimizes" by solving the resulting approximate optimization problem. 
For discrete decision spaces with a small number of alternatives, commonly known as ranking and selection with covariates, the offline stage focuses on learning the mapping from covariates to objective values for each alternative. Then, in the online stage, the objective values are predicted based on the newly observed covariate, and the best alternative is selected accordingly \citep{shen2021ranking, li2023convergence}. In contrast, when the decision space is continuous, directly evaluating the objective function across the entire feasible region is computationally prohibitive. To address this, surrogate models are constructed offline to approximate the objective function $\hat{f}\left(\theta; x\right)$, either locally or globally over the joint space of decisions $\theta$ and covariates $x$ \citep{hong2021surrogate}. Once a new covariate is observed, the surrogate model yields an approximated objective function that can be efficiently optimized using a deterministic solver. In both discrete and continuous cases, smoothing techniques such as linear regression and Gaussian process regression are commonly used to model the mapping from covariates to objective values.

The second stream follows the OTP approach. Here, the objective function is first ``optimized" at a representative set of covariate values during the offline stage, and the resulting covariate–optimizer pairs are used to learn a mapping from covariates to optimal solutions. In the online stage, once a new covariate is observed, the associated optimizer can be quickly ``predicted" using this learned mapping. For ranking and selection with covariates, where there are only a finite number of alternatives, \cite{keslin2022classification, keslin2025ranking} were the first to study the OTP approach. They formulate the problem into a classification problem, in which the covariate space is partitioned into subregions, each associated with an optimal alternative, and establish corresponding statistical guarantees. For continuous simulation optimization problems, existing work in this stream typically assumes that $f(\theta; x)$ is strongly convex in $\theta$ for every $x$, which ensures the uniqueness of the optimizer and thus a well-defined covariate-to-optimizer mapping. This optimal-solution function is then learned using various smoothing techniques, including neural networks \citep{zhang2021neural} and $k$-nearest neighbors \citep{luo2024reliable}.

Given the CSCSO setting considered in this work, we adopt the OTP approach to achieve efficiency in both optimization and prediction. From an optimization perspective, the OTP approach fully exploits the strong convexity of the objective function, allowing SO algorithms to attain fast convergence rates. Moreover, since all optimizations are conducted offline, it benefits from ample computational resources and is free from real-time constraints. In contrast, the PTO approach constructs surrogate models that may not preserve strong convexity, potentially resulting in only local optimal solutions \citep{bertsimas2022data}. Additionally, the PTO approach requires solving an optimization problem online, introducing latency in time-sensitive decision-making scenarios, particularly in resource-constrained environments such as on-device or edge computing. From a prediction perspective, the OTP approach learns a mapping from $\mathcal{X}\subset\mathbb{R}^{d}$ to $\Theta\subset\mathbb{R}^q$. In contrast, the PTO approach attempts to learn over $\mathcal{X}\times\Theta\subset\mathbb{R}^{q+d}$, which is more susceptible to the curse of dimensionality, especially when $q$ is large. Notably, the OTP approach can further leverage the smoothness of the optimal solution mapping, a property we formally establish in Section~\ref{sec:alg}, to enhance both the accuracy and efficiency of prediction through smoothing techniques.

Despite its practical appeal, the OTP approach remains theoretically underexplored. In particular, the ``optimization'' stage is often idealized in existing studies, which typically assume access to the ``exact'' optimal solution at each covariate, free from bias and variance. 
In practice, however, solving SO problems involves stochastic algorithms, which are inherently subject to sampling noise and optimization error, especially under finite simulation budgets. 
This ``inexactness'' fundamentally alters the design principles of both the prediction mechanism and the sample-allocation rule.

From the perspective of prediction, when exact solutions are available, the task reduces to interpolation. Under this idealized setting, \citet{luo2024reliable} establish convergence rates for the optimality gap using k-nearest neighbors, yielding a rate of $q^2n^{-2/d}$.  \citet{zhang2021neural} employ neural networks for prediction but do not provide a theoretical analysis of its performance. In contrast, traditional prediction with noisy data typically falls under the framework of regression, which focuses on unbiased observations with finite variance \citep{james2013introduction}. However, this framework fails to capture the unique characteristics of the OTP approach, where the inexact solutions obtained from finite-sample stochastic algorithms exhibit both bias and variance.  As a result, a rigorous convergence-rate analysis for prediction under such inexactness remains lacking.

From the perspective of sample allocation, existing studies typically focus only on the sample size of covariates. In the OTP setting, however, one must also account for the simulation effort required to solve the SO problem at each covariate. 
For example, suppose that stochastic gradient descent (SGD) is used to solve SO problems and each problem is allowed $T$ SGD iterations. As each SGD iteration consumes one simulation observation, then $T$ observations are required for each covariate. Consequently, the total sampling effort becomes proportional to the number of the covariates multiplied by the sampling effort per each covariate, making the trade-off between these two components a central consideration in the design of efficient OTP algorithms with inexact solutions.

This shift in the underlying philosophy of prediction and sample allocation gives rise to several fundamental questions. (1) \textit{Do the predicted optimal solutions and their corresponding objective values converge?} (2) If so, under a limited sampling budget, \textit{how ``exact'' must the optimization be at each covariate to achieve the optimal convergence rate?} In other words, is it necessary to solve each SO problem to high precision, effectively reducing the setting to interpolation? Or, alternatively, can a larger number of inexact solutions across a broader range of covariates lead to better predictive efficiency? These questions define the core theoretical challenges of the framework: namely, the {\it convergence-rate analysis} and the {\it optimal sample-allocation rule} for the OTP algorithms with inexact solutions.

To address the aforementioned theoretical challenges, we develop a comprehensive framework that captures the effect of inexact solutions in the OTP setting and examines their impact on convergence behavior and sample allocation. Our main contributions are summarized as follows:

First, we develop a unified algorithmic and theoretical framework for the OTP approach.  
On the algorithmic side, we formally exploit strong convexity to establish the well-definedness and smoothness of the optimal-solution function, which provide the foundation for function approximation within this approach.  
On the theoretical side, we adopt the optimality gap as the performance measure and explicitly account for the impact of inexact solutions.  
We establish convergence rates in terms of the optimality gap, derive optimal sample-allocation rules between the number of design covariate points, \(n\), and the optimization effort per covariate, \(T\), under a fixed computational budget \(\Gamma=nT\), and characterize the corresponding optimal convergence rates.

Second, we use Polyak–Ruppert averaging SGD (PR-SGD) algorithm \citep{polyak1992acceleration, Ruppert1991} as an illustrative example to quantify the inexactness introduced by SO algorithms. We choose the PR-SGD algorithm because it promotes more stable convergence by averaging iterates across steps \citep{nemirovski2009robust}. We rigorously characterize the bias and variance of the resulting solutions, with particular emphasis on establishing the order of the bias, which has not been well studied in the existing literature.

Third, we analyze several representative smoothing techniques, k-nearest neighbors (kNN), kernel smoothing (KS), linear regression (LR) and kernel ridge regression (KRR), within the proposed framework when the PR-SGD algorithm is used to solve SO problems. This leads to several noteworthy theoretical findings: 
(1) Although the optimal sample-allocation rule differs across techniques, all smoothing techniques require $T$ to be sufficiently large to ensure a desirable level of solution exactness. However, increasing $T$ does not always lead to better performance, especially for nonparametric techniques in high-dimensional settings. 
(2) In terms of convergence rates, we show that all of these smoothing techniques can recover the optimal rates observed in classical regression, despite the presence of optimization-induced bias.
(3) In particular, KRR demonstrates clear advantages in high-dimensional settings by effectively leveraging smoothness to mitigate the curse of dimensionality that severely limits nonparametric approaches such as kNN and KS.

The remainder of the paper is organized as follows. 
Section~\ref{sec:alg} establishes the well-definedness and smoothness of the optimal-solution function and introduces a unified algorithmic framework based on the OTP approach.  
Section~\ref{sec:inexact} develops a framework for analyzing the performance of the OTP algorithms by explicitly accounting for solution inexactness, and uses the PR-SGD algorithm as an illustrative example to quantify the solution inexactness.  
Section~\ref{sec:analysis} examines four representative smoothing techniques and derives their optimal sample-allocation rules and convergence rates when the PR-SGD algorithm is used. 
This section also compares the OTP approach with classical regression to provide further insights into the role of inexactness and discusses practical guidance on selecting smoothing techniques.  
Section~\ref{sec:num} presents numerical experiments that validate the theoretical results and demonstrate the effectiveness of the OTP approach.  
The paper is concluded in Section~\ref{sec:conclusion}, with additional proofs and numerical details in the Appendix ~\ref{app:smoothness}–\ref{app:num}.

In the analysis throughout this work, we use the notation ``$\asymp$'' to denote asymptotic equivalence: for two positive sequences $\{a_n\}$ and $\{d_n\}$,
we write $d_n \asymp a_n$ if there exist $C,C'>0$ such that $C a_n \le d_n \le C' a_n$, as $n \to \infty$.
Similarly, we write $d_n \gtrsim a_n$ if $d_n \ge C a_n$ for some $C>0$, and $d_n \lesssim a_n$ or $d_n=O\left(a_n\right)$ if $d_n \le C' a_n$ for some $C'>0$.
For a sequence of random variables $\{X_n\}$, we write $X_n = O_{\rm P}\left(a_n\right)$ if for every $\delta > 0$ there exists $M > 0$ such that $\Pr\!\big(|X_n| > M a_n\big) < \delta$, for all sufficiently large $n$.
The soft-$O$ and soft-$O_{\rm P}$ notations, $\tilde{O}\left( \cdot \right)$ and $\tilde{O}_{\rm P}\left( \cdot \right)$, are variants that ignore logarithmic factors. For example, $d_n = \tilde{O}\left(a_n\right)$ denotes $a_n = O\!\left(d_n \log^\kappa d_n\right)$ for some constant $\kappa>0$.
We also use the little-$o$ notation:
we write $d_n = o\left(a_n\right)$ if $d_n/a_n \to 0$ as $n \to \infty$.

\section{The ``Optimize then Predict'' Algorithmic Framework }\label{sec:alg}

In this section, we first establish the theoretical foundation of the OTP approach by demonstrating the well-definedness and smoothness of the optimal-solution function.
We then present a unified algorithmic framework based on the OTP approach, which is capable of addressing the challenges of real-time CSCSO and accommodating a broad class of optimization algorithms and smoothing techniques.

\subsection{The Optimal-Solution Function}
We denote the optimal-solution function as the mapping from the covariates to the optimizer, $\theta_*:\mathcal{X}\to\Theta$, and the optimal solution conditioned on $x$ is $\theta_*\!\left(x\right)=\left(\theta_*^1\left( x \right),\ldots,\theta_*^q\left( x \right)\right)$.
The OTP approach relies on constructing an approximation of $\theta_*\!\left(x\right)$, which intrinsically requires that $\theta_*\!\left(x\right)$ is well-defined and possesses sufficient smoothness to facilitate the application of smoothing techniques for prediction.
To show that these requirements are typically satisfied, we first make the following two assumptions.
% In this subsection, we show that under strong convexity and the smoothness condition on the objective function $f\left(\theta; x\right)$, specified in Assumption~\ref{as:smoothness}, and the  interiority condition in Assumption~\ref{as:interior}, $\theta_*$ is indeed well-defined and smooth.
% These properties are formally established in Proposition~\ref{prop:smoothness}.
\begin{assumption}\label{as:smoothness}
The function $f\left(\theta; x\right)$ is $C^{m+1}$-smooth  in $\left(\theta,x \right)\in \Theta\times \mathcal{X}$ for some $m\geq 1$, where both \(\Theta\) and \(\mathcal{X}\) are compact sets. 
Moreover, for every $x\in\mathcal{X}$, the function $f(\theta;x)$ is strongly convex in $\theta$. 
% where $\Theta\subset \mathbb{R}^q$ and $\mathcal{X}\subset \mathbb{R}^d$. 

\end{assumption}
Assumption \ref{as:smoothness} imposes some regularity conditions on the objective function $f\left(\theta; x\right)$.
The smoothness condition implies that, for any $|\alpha|+|\beta|\leq m+1$, the mixed partial derivatives $D^\alpha D^\beta f\left(\theta; x\right)$ exist and are continuous, where
\[
D^\alpha D^\beta f\left(\theta; x\right)
=\frac{\partial^{|\alpha|+|\beta|}f\left(\theta; x\right)}{\partial (\theta^1)^{\alpha_1}\ldots\partial(\theta^q)^{\alpha_q} \partial (x^1)^{\beta_1}\ldots\partial (x^d)^{\beta_d}},
\]
and $\alpha=\left(\alpha_1,\ldots,\alpha_q\right)$, $\beta = \left( \beta_1 , \ldots , \beta_d \right)$ are the vectors of non-negative integers with $|\alpha|=\alpha_1+\ldots+\alpha_q$ and $|\beta|=\beta_1+\ldots+\beta_d$. The assumption $m\geq 1$, i.e., the objective function is at least twice continuously differentiable with respect to $\theta$, is commonly adopted and plays a crucial role in establishing the theoretical guarantees of SO algorithms \citep{polyak1992acceleration}.

\begin{assumption}\label{as:interior}
The optimal solution $\theta_*\!\left(x\right)$ lies in the interior of $\Theta$ for every $x\in\mathcal{X}$.
\end{assumption}
Assumption \ref{as:interior} is a conventional assumption in the field of stochastic approximation \citep{robbins1951stochastic,polyak1992acceleration, cutler2024stochastic}. It implies that $\nabla f(\theta_*\!\left(x\right);x)=0$ for every $x\in\mathcal{X}$.

Given the above two assumptions, we have the following proposition that ensure the well-definedness and smoothness of the optimal-solution function $\theta_*\!\left(x\right)$.

\begin{proposition}\label{prop:smoothness}
Under Assumption~\ref{as:smoothness} and \ref{as:interior},  $\theta_*:\mathcal{X}\to \Theta$ 
is well-defined (i.e., single-valued), and belongs to $C^{m}(\mathcal{X})$.
\end{proposition}

The strong convexity plays a crucial role in establishing Proposition~\ref{prop:smoothness}.
First, it implies that, for any covariate $x\in\mathcal{X}$, the objective function $f\left(\theta; x\right)$ admits a unique minimizer $\theta_*\!\left(x\right)$, therefore ensures the existence of the well-defined mapping $\theta_*:\mathcal{X}\to\Theta$.
% On one hand, it ensures the existence of the one-to-one mapping $\theta_*:\mathcal{X}\to\Theta$. Specifically, the strong convexity of $f\left(\theta; x\right)$ w.r.t $\theta$ implies that, for any covariate $x\in\mathcal{X}$, the objective function $f\left(\theta; x\right)$ admits a unique minimizer $\theta_*\!\left(x\right)$. 
Second, by noticing that $\theta_*$ is the unique solution of $\nabla f(\theta_*\!\left(x\right);x)=0$, the strong convexity allows us to apply the implicit function theorem \citep{krantz2002implicit} to establish the smoothness of $\theta_*$ as a consequence of that of $f$. 
% On the other hand, it guarantees that $\theta_*$ inherits the smoothness of that of the objective function $f$. 
% Notice that $\theta_*$ is actually a implicit function such that $\nabla f(\theta_*\!\left(x\right);x)=0$.
% The strong convexity ensures that the Jacobian of this system, namely the Hessian matrix, i.e., the Hessian matrix $\nabla^2_{\theta\theta}f(\theta_*\!\left(x\right);x)$, is positive definite, and hence invertible at every $x\in\mathcal{X}$.
% This allows us to apply the implicit function theorem \citep{krantz2002implicit} to establish the smoothness of $\theta_*$ as a consequence of that of $f$. 
The detailed proof of Proposition \ref{prop:smoothness} is provided in Appendix \ref{app:smoothness}.

Proposition~\ref{prop:smoothness} constitutes a critical foundation for the ``predict'' stage of the OTP algorithms.
First, the well-defined mapping $\theta_*\!\left(x\right)$ provides the validity of the OTP approach, which is based on function  approximation of optimal-solution mapping.
In contrast, if multiple optimal solutions exist for a given covariate, the mapping from covariates to optimizers becomes ambiguous, precluding the application of standard function approximation techniques.
Second, the smoothness of $\theta_*\!\left(x\right)$ is essential for effective function approximation. Most smoothing techniques require a certain degree of smoothness in the target function, especially for non-parametric techniques in high-dimensional settings  \citep{Tsybakov2009}.
% Therefore, strong convexity serves as an essential foundation of the OTP approach. In the ``optimize'' stage, it guarantees faster convergence of simulation optimization  (reference). In the ``predict'' stage, it provides the theoretical justification for function approximation by ensuring the well-definedness and smoothness of the optimal-solution function, as established in in Proposition~\ref{prop:smoothness}.

\subsection{The Algorithmic Framework}\label{subsec:alg}
After establishing the theoretical foundation of the OTP approach, we now summarize the main procedure of the unified algorithmic framework tailored to the real-time CSCSO problem. The framework consists of two stages. In the offline stage, we first solve the SO problems for a set of covariates selected based on a quasi-uniform design, using a chosen optimization algorithm with a given budget. We then  collect the covariate-solution pairs and apply a smoothing technique to approximate the optimal-solution function with respect to the covariates. In the online stage, once the value of the covariate is observed, we plug it into the optimal-solution function to obtain the approximated optimal solution. By shifting the computationally intensive simulation and optimization to the offline stage, this framework enables fast online decision-making. Moreover, it maintains flexibility in selecting both the algorithm for SO and the smoothing technique for prediction. The complete procedure is summarized in Algorithm \ref{alg}.

\begin{algorithm}[ht]
\caption{The ``Optimize then Predict'' Algorithmic Framework}\label{alg}
\hspace{2pt}
\begin{algorithmic}
\State \textbf{The Offline Stage: Optimization and Function Approximation}

\State $-$ Select $n$ covariates $\{x_i\}_{i=1}^n$ based on a quasi-uniform design on $\mathcal{X}$.

\State $-$ For $i=1,2,\ldots,n$, employ an optimization algorithm to solve $\min_{\theta\in\Theta}f(\theta;x_i)$ with a budget of $T$ simulation experiments, and denote the solution as $\bar\theta_T\left(x_i\right)=\left(\bar\theta_T^1\left(x_i\right),\ldots,\bar\theta_T^q\left(x_i\right)\right)$. 

\State $-$ Fit the optimal-solution function $\theta_*\left( \cdot \right)$ using $\left\{\left(x_{i},\bar\theta_T\left(x_i\right)\right)\right\}_{i=1}^{n}$  by a smoothing technique. Let $\hat{\theta}\left( \cdot \right)$ denote the fitted solution function.

\vspace{6pt}

\State \textbf{The Online Stage: Prediction}

\State $-$ After observing the value of the covariate, say $x=x'$, report $\hat{\theta}(x')=\left(\hat{\theta}^1(x'),\ldots, \hat{\theta}^q(x')\right)$ as the predicted optimal solution to  $\min_{\theta\in\Theta}f(\theta,x')$.
\end{algorithmic}

\end{algorithm}

There are a few remarks that we want to make about this algorithm. 
First, we adopt a quasi-uniform approach \citep{wendland2004scattered} to design the set of covariates $D=\{x_i\}_{i=1}^n$ in the offline stage. 
Let $h_n=\sup_{x\in\mathcal{X}}\min_{1\leq i \leq n}\|x-x_i\|$ be the  fill distance, which reflects the coverage of the design points; a smaller value indicates better coverage and reduces the worst-case prediction error.
Let $q_n = {1\over2} \min_{ i\neq j}\|x_i-x_j\|$ be the separation distance, which reflects the uniform spacing of the design points; a larger value prevents clustering and ensures that the points are well-dispersed.
The quasi-uniform design requires the fill distance and separation distance are of the same order, $n^{-1/d}$, i.e., $q_n \asymp h_n\asymp n^{-1/d}$.
Such quasi-uniform design ensures the sufficient exploration of the covariate space, enhancing prediction performance.
Typical quasi-uniform constructions include fixed grid designs, greedy packing, energy minimization and so on \citep{pronzato2023quasi}.

Second, for each covariate $x_i$, $i=1,\ldots,n$, we employ an optimization algorithm to solve the SO problem  $\min_{\theta\in\Theta}f(\theta,x_i)$. Unlike existing OTP algorithms in the literature (e.g., \cite{luo2024reliable}), we do not assume access to the exact optimal solutions $\theta_*\left(x_i\right)$. Instead, we allocate a fixed budget of $T$ simulation experiments per covariate to obtain an approximate solution $\bar\theta_T\left(x_i\right)$, whose accuracy is controlled by $T$.
A central focus of this work is to understand how the inexactness of $\bar\theta_T\left(x_i\right)$ affects the overall algorithm performance, and how to appropriately select $T$ under a total simulation budget of $\Gamma$. In our framework, the total budget satisfies $\Gamma = nT$, revealing a fundamental trade-off between the number of design points in the covariate space and the accuracy of the corresponding optimized solutions.

Third, in this paper, we consider the smoothing techniques that admit the representation
\begin{equation}\label{eq:representation}
\hat{\theta}\!\left(x\right)=\sum_{i=1}^n w\!\left(x_i,\, x\right)\,\bar{\theta}_T\left(x_i\right),   
\end{equation}
where $w\!\left(x_i,\, x\right)$ denotes the smoothing weight assigned to $\bar{\theta}_T\left(x_i\right)$.
This representation is particularly convenient for our subsequent theoretical analysis and encompasses a broad class of smoothing techniques, including parametric models such as linear regression and non-parametric models such as kNN, KS and KRR. As examples, we consider both linear regression and kNN. For linear regression, the weight function is $w\!\left(x_i,\, x\right)=\langle\phi\left(x_i\right),\phi\left( x \right)\rangle$ with orthonormal basis functions \(\phi\left( x \right) = (\phi_1\left( x \right), \dots, \phi_s\left( x \right))^\top \in \mathbb{R}^s\), and for kNN, the weight function is $w\!\left(x_i,\, x\right)=1/k$ if $x_i$ is one of the $k$ nearest neighbors of $x$ and 0 otherwise.

Equation~\eqref{eq:representation} implies that  $\hat{\theta}^j\left( x \right) = \sum_{i=1}^n w\!\left(x_i,\, x\right)\,\bar{\theta}_T^j\left(x_i\right)$ for each dimension $j=1,\ldots,q$.
Therefore, applying the smoothing technique to the vector \(\bar{\theta}_T\left(x_i\right)\) is equivalent to applying it componentwise to each coordinate \(\bar{\theta}_T^j\left(x_i\right)\), as if approximating \(q\) separate functions. 
From a theoretical standpoint, this componentwise perspective allows us to leverage the extensive body of theoretical results for smoothing techniques, which are typically developed for scalar outputs. 
From a computational standpoint, the procedure effectively approximates a single function rather than \(q\) separate functions, as the weights \(w\!\left(x_i,\, x\right)\) are shared across all $q$ dimensions.  
For example, in LR once \(w\!\left(x_i,\, x\right)=\langle \phi\left(x_i\right),\phi\left( x \right)\rangle\) is computed, the same weights can be applied simultaneously to all \(q\) coordinates, i.e., \(\sum_{i=1}^n \langle \phi\left(x_i\right),\phi\left( x \right)\rangle\,\bar{\theta}_T^j\left(x_i\right),\ j=1,\ldots,q\), without additional overhead.

\subsection{The Optimality Gap}\label{subsec:opt_gap}
To evaluate the performance of the predicted solution $\hat{\theta}\!\left(x\right)$, we adopt the optimality gap $f\left(\hat{\theta}\!\left(x\right);x\right)-f\left(\theta_*\!\left(x\right);x\right)$ at the target covariate $x$ as the performance metric of the algorithmic framework.
As the predicted solution $\hat{\theta}\!\left(x\right)$ depends on the solutions $\bar{\theta}_T\left(x_i\right), i=1,\ldots,n$,  obtained through SO algorithms with a finite budget, $\hat{\theta}\!\left(x\right)$ is random. Therefore, we first consider the expected optimality gap $\oE\left[f\left(\hat{\theta}\!\left(x\right);x\right)-f(\theta_*\!\left(x\right);x)\right]$.
In the following proposition, we show that it has the same order as the mean squared error (MSE) of the predicted solution $\hat{\theta}\!\left(x\right)$.
It is based on the second-order Taylor's expansion of the objective function at $\theta_*\!\left(x\right)$, where the first-order term is zero because $\nabla f\left(\theta_*\!\left(x\right);x\right)=0$ by Assumption~\ref{as:interior}.
The detailed proof can be found in Appendix~\ref{app:prop_opt_gap}.
% We start from the analysis of expected optimality gap $\oE\left[f(\hat{\theta}\!\left(x\right);x)\right]-f(\theta_*\!\left(x\right);x)$, where the expectation is taken with respect to the randomness from the stochastic optimization, as reflected in the offline solutions $\bar{\theta}_T\left(x_i\right)$ conditioned on the covariate $x_i$ for $i=1,\ldots,n$. 
% Under Assumption~\ref{as:interior}, we apply Taylor's expansion to establish the direct connection between the expected optimality gap and the MSE of predicted solution $\hat{\theta}\!\left(x\right)$, as summarized in the following proposition.
\begin{proposition}\label{prop:opt_gap}
Under Assumption~\ref{as:smoothness}, we have $\oE\left[f\left(\hat{\theta}\!\left(x\right);x\right)-f\left(\theta_*\!\left(x\right);x\right)\right]=O\left(\oE\left[\left\|\hat\theta\!\left(x\right)-\theta_*\!\left(x\right)\right\|^2\right]\right)$, where $\|\cdot\|$ denote the $L_2$ norm. 
\end{proposition}

We further derive the convergence rate of the optimality gap in terms of the order of the MSE of $\hat{\theta}\!\left(x\right)$ as follows.
\begin{corollary}\label{col:convergence}
Suppose that $\oE\left[\left\|\hat\theta\!\left(x\right)-\theta_*\!\left(x\right)\right\|^2\right]\to 0$ when $n,T\to\infty$.
Then, the optimality gap $f\left(\hat{\theta}\!\left(x\right);x\right)- f\left(\theta\!\left(x\right);x\right)\to 0$ in probability and converges at the rate of $O_{\rm P}\left(\oE\left[\left\|\hat\theta\!\left(x\right)-\theta_*\!\left(x\right)\right\|^2\right]\right)$.
\end{corollary}

Corollary~\ref{col:convergence} follows directly from Proposition~\ref{prop:opt_gap} by Markov’s inequality (see the proof in Appendix~\ref{app:col_convergence}). 
First, it shows that the convergence of the MSE of the predicted solution, i.e., $\oE\left[\left\|\hat\theta\!\left(x\right)-\theta_*\!\left(x\right)\right\|^2\right]$, is sufficient to guarantee convergence of the algorithm in terms of the optimality gap. 
Furthermore, 
% along with Proposition~\ref{prop:opt_gap}, 
it shows that the optimality gap converges at the same rate as $\oE\left[\left\|\hat\theta\!\left(x\right)-\theta_*\!\left(x\right)\right\|^2\right]$.
In the rest of the paper, we focus on the asymptotic behavior of $\oE\left[\left\|\hat\theta\!\left(x\right)-\theta_*\!\left(x\right)\right\|^2\right]$, as it is directly linked to the asymptotic behavior of smoothing techniques and, therefore, easier to analyze.
% the convergence rate of the optimality gap is characterized explicitly by the order of the MSE of $\hat{\theta}\!\left(x\right)$, which motivates our subsequent analysis of the MSE.

\section{The Impact of Inexact Solutions}\label{sec:inexact}

As we establish in Section~\ref{sec:alg}, the optimality gap of the predicted solution of Algorithm~\ref{alg}, i.e., $f\left(\hat{\theta}\!\left(x\right);x\right)- f\left(\theta\!\left(x\right);x\right)$, converges at the same rate as $\oE\left[\left\|\hat\theta\!\left(x\right)-\theta_*\!\left(x\right)\right\|^2\right]$. To understand how inexact solutions $\bar{\theta}_T\left(x_1\right),\ldots,\bar{\theta}_T\left( x_n \right)$ impact the optimality gap, we first provide the following proposition that offers a decomposition of the MSE.
The proof of the proposition is included in Appendix~\ref{app:prop_MSE_decom}.

\begin{proposition}\label{prop:MSE_decom}
Suppose that Assumptions~\ref{as:smoothness} and \ref{as:interior} are satisfied, Algorithm~\ref{alg} is used to solve problem~\eqref{pr:CSCSO}, and the smoothing technique used in algorithm admits the representation~\eqref{eq:representation}. 
Then, 
\begin{equation}\label{eq:dim_decom}
\oE\left[\left\|\hat\theta\!\left(x\right)-\theta_*\!\left(x\right)\right\|^2\right]
%=\oE\left[\sum_{j=1}^q\left(\hat\theta^{j}\left( x \right)-\theta_*^{j}\left( x \right)\right)^2\right]
=\sum_{j=1}^q\oE\left[\left(\hat\theta^j\left( x \right)-\theta_*^{j}\left( x \right)\right)^2\right], 
\end{equation}
where
\begin{align}
 &\oE\left[\left(\hat\theta^j\left( x \right)-\theta_*^{j}\left( x \right)\right)^2\right] \nonumber\\
&\quad =
\Bigg[\underbrace{\sum_{i=1}^nw\!\left(x_i,\, x\right)\theta_*^j\left(x_i\right) - \theta_*^j\left( x \right)}_{\text{interpolation bias}}
+\sum_{i=1}^nw\!\left(x_i,\, x\right)\underbrace{\left(\oE\left[\bar{\theta}_T^j\left(x_i\right)\right]-\theta_*^j\left(x_i\right)\right)}_{\text{solution bias}}\Bigg]^2
+\sum_{i=1}^nw^2\left(x_i, x\right)\underbrace{\oV\left[\bar{\theta}_T^j\left(x_i\right)\right]}_{\text{solution variance}}.\label{eq:MSE_decom}
\end{align}  
\end{proposition}
\vspace{6pt}

There are two important implications from Proposition~\ref{prop:MSE_decom}.
First, notice that the componentwise MSEs are of the same order across \(j\), since the weights \(w\!\left(x_i,\, x\right)\) are identical across all components and the solution bias and variance exhibit homogeneous orders across dimensions.  
As a consequence,  Proposition~\ref{prop:MSE_decom} implies that
$\oE\left[\left\|\hat{\theta}\!\left(x\right)-\theta_*\!\left(x\right)\right\|^2\right] \asymp q \,\oE\left[\left(\hat{\theta}^j\left( x \right)-\theta_*^j\left( x \right)\right)^2\right]. $ 
Therefore, we only need to analyze the convergence rate of the MSE of a single dimension to know that of the of the overall MSE.
It also implies that the overall MSE as well as the optimality gap, i.e., $f\left(\hat{\theta}\!\left(x\right);x\right)- f\left(\theta\!\left(x\right);x\right)$, have a convergence rate that grows linearly in $q$, which is not immune to the dimension of decision variable, but certainly avoids the typical exponential type of curse of dimensionality often seen in predict-then-optimize algorithms \citep{fan2025review}.

Second, Proposition~\ref{prop:MSE_decom} shows that the component MSE can be decomposed into three error terms. 
The first one is related to the interpolation error of exact solutions, the second and the third terms are determined by the biases and variances of the inexact solutions, respectively.
Previous literature on OTP typically assumes the access to the exact solutions. Therefore, it overlooks the second and third terms and thus results in overly optimistic convergence rates.
In classical regression analysis, the solutions are typically assumed unbiased, effectively setting the second term to zero.
Therefore, its convergence rate is not directly applicable to our problem either. In our analysis, we need to explicitly account for all three error terms, which better reflect the reality in practical applications.

\subsection{Inexact Solution from Polyak–Ruppert Averaging SGD Algorithm} 
\label{subsec:PR_SGD}

Notice that the inexactness of the solution, i.e., its bias and variance, depends on the specific SO algorithm employed in the offline stage of Algorithm~\ref{alg}.
In this subsection, we use PR-SGD algorithm as an illustrative example to demonstrate the solution bias and variance, because the SGD algorithms are widely used and the PR-SGD variant of SGD has stable solutions that are desirable in our applications. 
We emphasize, however, that the proposed algorithmic and theoretical framework may be applied to other SO algorithms as well.

We first briefly review the PR-SGD algorithm under the framework of CSO, which takes the following iterative scheme for a given covariate $x$: 
\begin{eqnarray*}
    \theta_0\left( x \right) &\in &  \Theta,\\
    \theta_t\left( x \right) &=& \Pi_\Theta\left[\theta_{t-1}\left( x \right)-\gamma_t \nabla_\theta F\left(\theta_{t-1}\left( x \right);x\right)\right], \quad \text{for}\ t=1,\ldots,T, \\
    \bar{\theta}_T\left( x \right) &=& {1\over T+1}\sum_{t = 0}^{T}\theta_t\left( x \right), 
\end{eqnarray*}
where $\theta_0\left( x \right)$ is the initial solution in the feasible region $\Theta$, $\nabla_\theta F\left(\theta; x\right)$ is an unbiased gradient estimator of $\nabla_\theta f\left(\theta; x\right)$ obtained from a single simulation experiment, $\gamma_t$ is the deterministic step size at iteration $t$, $\Pi_\Theta[\cdot]$ is the projection that brings a solution outside back to the feasible region $\Theta$, and $\bar{\theta}_T\left( x \right)$ is the reported solution after the algorithm completes all $T$ iterations. The averaging operation over multiple steps can mitigate fluctuations caused by noisy gradients and promote more stable convergence \citep{nemirovski2009robust, kim2014guide}.

Based on the technical analysis of \cite{polyak1992acceleration}, with the additional regularity conditions used by them, see Assumptions~\ref{as:3.1}-\ref{as:3.4} in Appendix~\ref{app:bias_conditions}, we have the following proposition on the bias and variance of the solution $\bar{\theta}_T\left(x_i\right)$.
Notice that \citet{polyak1992acceleration} establish the asymptotic normality of $\bar\theta_T\left( x \right)$, which directly implies the variance is of order $\mathbb{V}\left[\bar\theta_T^j\left( x \right)\right]=O\left(1/T\right)$ for $j=1,\ldots,q$, but the bias is  of order $o\left( 1/\sqrt T\right)$.  
We prove explicitly that the bias is of $\tilde{O}\left(1/T\right)$ under the same set of conditions.
The detailed proof is included in Appendix~\ref{app:bias_proof}.
\begin{proposition}\label{prop:bias}
Under Assumptions \ref{as:smoothness}-\ref{as:3.4}, for the solution obtained from the $T$-step PR-SGD algorithm, we have $\oE\left[\bar\theta_T^j\left( x \right)\right]-\theta_*^j\left( x \right)=\tilde O\left(1/T\right)$ and $\oV\left[\bar\theta_T^j\left( x \right)\right]=O\left(1/T\right),\ j=1,\ldots,q$.
\end{proposition}

In the rest of the paper, we explicitly assume that the PR-SGD algorithm is used in the offline stage to solve SO problems and derive the convergence rate and optimal sample-allocation rules accordingly.
However, we want to emphasize that the analysis procedure adopted in the paper may be easily extended to other SO algorithms as well.

\section{Optimal Sample Allocation and Convergence Rate}\label{sec:analysis}

In Section~\ref{subsec:PR_SGD}, we establish the orders of solution bias and variance for the PR-SGD algorithm.  
To derive the explicit order of $\oE\left[\left(\hat\theta^j\left( x \right)-\theta_*^{j}\left( x \right)\right)^2\right]$, which is directly linked to the optimality gap $f\left(\hat{\theta}\!\left(x\right);x\right)- f(\theta\!\left(x\right);x)$, it is also necessary to specify the order of the interpolation bias and the representation of the weights $w\!\left(x_i,\, x\right)$, according to the MSE decomposition in Equation~\eqref{eq:MSE_decom}.  
Since both depend on the smoothing technique applied, in this section we examine four representative techniques: two classical nonparametric techniques, k-nearest neighbors (kNN) and kernel smoothing (KS); a parametric technique, linear regression (LR); and an advanced machine learning technique, kernel ridge regression (KRR).  
For each smoothing technique, we first derive the componentwise MSE of the predicted solution \(\hat{\theta}^j\left( x \right)\), given \(n\), \(T\), and the hyperparameters.  
We then establish the optimal sample-allocation rule, the associated optimal order of hyperparameters, and the resulting convergence rate.  
In particular, we elaborate more on the kNN technique first to establish the analysis framework and highlight the key points of other techniques subsequently.

% Theorems~\ref{thm:kNN}–\ref{thm:KRR} resolve the central theoretical challenge identified in Section~\ref{sec:intro}: establishing the convergence of the OTP framework under inexact solutions and determining the corresponding optimal sample allocation for the algorithm.  

\subsection{k-Nearest Neighbors}\label{subsec:knn}

kNN is a classical nonparametric technique for pointwise function estimation.
The kNN estimator $\theta_{kNN}\left( x \right)$ is defined as the average of the $\theta$-values of the $k$ training points whose inputs $x_i$'s are closest to $x$ in the covariate space, which can be formulated as: 
\begin{equation}\label{eq:kNN}
\hat\theta_{kNN}\left( x \right) = {1\over k}\sum_{i\in N_k\left( x \right)}\bar{\theta}_T\left(x_i\right),
% = {1\over k}\sum_{i\in N_k\left( x \right)}\left(\theta^*\left(x_i\right) + {\xi\left(x_i\right)\over\sqrt{T}} + {\varepsilon\left(x_i\right)\over T}\right),   
\end{equation}
where \(N_k\left( x \right)\) contains the indices of the \(k\) closest points of \(x_1, \ldots, x_n\) to \(x\), in terms of Euclidean distance.  
Therefore, $w\!\left(x_i,\, x\right)=1/k$ for $i\in N_k\left( x \right)$ while $w\!\left(x_i,\, x\right)=0$ for $i\notin N_k\left( x \right)$ in the kNN estimator.
To ensure the convergence, it is required that $k\to\infty$ and $k/n\to 0$ as $n\to\infty$ \citep{gyorfi2006distribution}.
Based on Equation~\eqref{eq:kNN}, we derive the order of $\oE\left[\left(\hat\theta^j_{kNN}\left( x \right)-\theta_*^{j}\left( x \right)\right)^2\right]$ and summarize it in the following lemma.

\begin{lemma}[kNN]\label{lem:kNN}
Suppose Assumptions~\ref{as:smoothness}–\ref{as:3.4} hold.
Consider Algorithm~\ref{alg}, in which the offline SO problems are solved via a $T$-step PR-SGD algorithm and the optimal-solution function is approximated by kNN based on the sample $\left\{x_i, \bar{\theta}_T\left(x_i\right)\right\}_{i=1}^n$.
% , where $\bar{\theta}_T\left(x_i\right)$'s are the offline solutions at the designed covariates.
Then, for every dimension $j=1,\ldots,q$, the MSE of predicted solution satisfies
\begin{equation}\label{eq:kNN_MSE}
\oE\left[\left(\hat{\theta}_{kNN}^j\left( x \right)-\theta^j_*\left( x \right)\right)^2\right] = \tilde{O}\left(\left({k\over n}\right)^{2/d}+{1\over T^2}+{1\over kT}\right), \ \text{ as }\ n \text{ and } T\to \infty.
% , \quad j=1,\ldots,q.
\end{equation}
\end{lemma}

\vspace{6pt}

\begin{proof}{Proof.}
According to Equation~\eqref{eq:kNN} and  the MSE decomposition in Proposition~\ref{prop:MSE_decom}, the bias is
\begin{eqnarray*}
\mathbb{E}\left[\hat{\theta}_{kNN}^j\left( x \right)\right]-\theta_*^j\left( x \right)
% &=&\frac{1}{k}\sum_{i\in N_k\left( x \right)}\left(\mathbb{E}[\bar{\theta}_T^j\left(x_i\right)]-\theta_*^j\left(x_i\right)\right) +\frac{1}{k}\sum_{i\in N_k\left( x \right)}\theta_*^j\left(x_i\right)-\theta_*^j\left( x \right)\\
&=&\underbrace{\frac{1}{k}\sum_{i\in N_k\left( x \right)}\left(\mathbb{E}[\bar{\theta}_T^j\left(x_i\right)]-\theta_*^j\left(x_i\right)\right)}_{\text{solution-induced bias}}
+\underbrace{\frac{1}{k}\sum_{i\in N_k\left( x \right)}\left(\theta_*^j\left(x_i\right)-\theta_*^j\left( x \right)\right)}_{\text{interpolation bias}},
\end{eqnarray*}
where the equality holds due to $\sum_{i\in N_k\left( x \right)}1/k =1$ and thus $\theta_*^j\left( x \right)=\frac{1}{k}\sum_{i\in N_k\left( x \right)}\theta_*^j\left( x \right)$.

By Proposition~\ref{prop:bias}, when the PR-SGD algorithm is used, each offline solution bias $\oE\left[\bar{\theta}^j_T\left(x_i\right)\right]-\theta_*^j\left(x_i\right)=\tilde{O}\left(1/T\right)$, and since $\sum_{i\in N_k\left( x \right)}1/k=1$, the solution-induced bias satisfies
\[{1\over k}\sum_{i\in N_k\left( x \right)}\left(\oE\left[\bar{\theta}^j_T\left(x_i\right)\right]-\theta_*^j\left(x_i\right)\right)=\tilde{O}\left(\frac{1}{T}\right).
\]
For the interpolation bias, Proposition~\ref{prop:smoothness} and the compactness of $\mathcal{X}$ ensure that $\theta_*^j\left( \cdot \right)$ is $L$-Lipschitz, i.e., 
$\left|\theta_*^j\left(x_i\right)-\theta_*^j\left( x \right)\right|\le L\|x_i-x\|$ for some constant $L>0$.  
By Lemma~\ref{lem:counting} and Remark~\ref{remark}, which are included in Appendix~\ref{app:smoothing}, the neighbor distance satisfies $\|x_i-x\|\asymp\left(  k/n \right)^{1/d}$ for $i\in N_k\left( x \right)$, under the quasi-uniform design of $\{x_i\}_{i=1}^n$. Hence, the interpolation bias satisfies
\[{1\over k}\sum_{i\in N_k\left( x \right)}\left(\theta_*^j\left(x_i\right)-\theta_*^j\left( x \right)\right)\leq {L\over k}\sum_{i\in N_k\left( x \right)}\|x_i-x\|\asymp\left(  k/n \right)^{1/d}.\]  
Thus, the overall bias is 
\[
\mathbb{E}\left[\hat{\theta}_{kNN}^j\left( x \right)\right]-\theta_*^j\left( x \right)
= \tilde{O}\left(\frac{1}{T}+\left(\frac{k}{n}\right)^{1/d}\right).
\]

For the variance,
\[
\oV\left[\hat{\theta}_{kNN}^j\left( x \right)\right]
=\oV\left[\frac{1}{k}\sum_{i\in N_k\left( x \right)}\bar{\theta}_T^j\left(x_i\right)\right]
=\frac{1}{k^2}\sum_{i\in N_k\left( x \right)}\oV\left[\bar{\theta}_T^j\left(x_i\right)\right]
=O\!\left(\frac{1}{kT}\right),
\]
where the last step again follows from Proposition~\ref{prop:bias}.  

Combining the bias and variance yields
\[
\oE\left[\left(\hat{\theta}_{kNN}^j\left( x \right)-\theta^j_*\left( x \right)\right)^2\right] = \tilde{O}\left(\left({k\over n}\right)^{2/d}+{1\over T^2}\right) + O\left({1\over kT}\right) = \tilde{O}\left(\left({k\over n}\right)^{2/d}+{1\over T^2}+{1\over kT}\right), \quad j=1,\ldots,q,
\]
which gives the stated result.
\hfill\Halmos
\end{proof}

\smallskip

% The proof of Lemma~\ref{lem:kNN} illustrates a concrete application of Proposition~\ref{prop:MSE_decom}.  
% The bias consists of (i) the solution-induced bias, of order $\tilde{O}\left(1/T\right)$, and (ii) the interpolation bias, of order $\left(  k/n \right)^{1/d}$.  
% The solution-induced bias coincides with the order of the offline solution bias because the kNN weights satisfy $\sum_{i=1}^nw\!\left(x_i,\, x\right)=\sum_{i\in N_k\left( x \right)}{1\over k}=1$.
% The interpolation bias stems from the discrepancy between the target point $x$ and its neighboring design points $\{x_i\}_{i\in N_k\left( x \right)}$, which is of order $\left(  k/n \right)^{1/d}$ by Lemma~\ref{lem:counting}.
% The variance is determined by the sum of squared weights, $\sum_{i=1}^n w^2\left( x,x_i\right)=\sum_{i\in N_k\left( x \right)}1/k^2=1/k$, together with the solution variance $O\left(1/T\right)$, yielding $O(1/(kT))$.
% Hence, the effect of solution inexactness enters explicitly through both the solution-induced bias and the variance components.

For a fixed computation budget $\Gamma$, the optimal convergence rate of the MSE can be obtained by solving the following optimization problem:
\begin{eqnarray*}
\min_{n,T,k}\ \left({k\over n}\right)^{2/d}+{1\over T^2}+{1\over kT}, \quad
 \text{subject to}\quad nT=\Gamma,\ k^{-1}=o(1), \ k=o\left(n\right).
\end{eqnarray*}

The optimal solution and the corresponding convergence rate are established in the following theorem.
\begin{theorem}[kNN]
    \label{thm:kNN}
Suppose Assumptions \ref{as:smoothness}-\ref{as:3.4} hold.
Consider Algorithm~\ref{alg}, in which the offline SO problems are solved via a $T$-step PR-SGD algorithm and the optimal-solution function is approximated by kNN.
Giving  a total simulation budget $\Gamma$, the optimal sample-allocation rule satisfies 
$n^*\asymp\Gamma^{d\over d+2}k^*$, and $T^*\asymp\Gamma^{2\over d+2}/k^*$,
with the hyperparameter $k^*$ chosen such that $k^*T^*\asymp\Gamma^{2\over d+2}$ and $(k^*)^{-1}=o(1)$. Moreover, the order of $T^*$ lies in the interval $\left.\left[\Gamma^{1\over d+2}, \Gamma^{2\over d+2}\right)\right.$ and the order of $n^*=\Gamma/T^*$ lies in the interval $\left.\left(\Gamma^{d\over d+2}, \Gamma^{d+1\over d+2}\right]\right.$.
Under this sample-allocation rule and hyperparameter setting, the optimality gap $f\left(\hat{\theta}_{kNN}\left( x \right);x\right)- f\left(\theta\!\left(x\right);x\right)\to 0$ in probability and converges in the rate of $\tilde{O}_{\rm P}\left(q\Gamma^{-{2\over d+2}}\right)$.
\end{theorem}

\begin{proof}{Proof.}
We analyze two cases.

\emph{Case 1: $T \gtrsim k$.} 
The dominant terms in Equation\eqref{eq:kNN_MSE} are $\left(  k/n \right)^{2/d}$ and $1/(kT)$. 
Balancing them under the budget constraint $nT=\Gamma$ gives
\[
T^*\asymp \Gamma^{2/\left(d+2\right)}/k^*, \quad 
n^*\asymp \Gamma^{d/\left(d+2\right)}k^*, \quad 
k^*T^*\asymp \Gamma^{2/\left(d+2\right)}, \quad 
(k^*)^{-1}=o(1).
\]
Since $T\gtrsim k$, we need $T^*\gtrsim \Gamma^{1/\left(d+2\right)}$ and thus $T^*\in\left[\Gamma^{1\over\left(d+2\right)},\Gamma^{2\over\left(d+2\right)}\right)$. 
Hence, $\oE\left[\left(\hat{\theta}_{kNN}^j\left( x \right)-\theta^j_*\left( x \right)\right)^2\right]=\tilde{O}\left(\Gamma^{-{2\over d+2}}\right)$.

% \emph{Case 1: $T \gtrsim k$.}  
% In this regime, the dominant terms in Equation~\eqref{eq:kNN_MSE}  are $\left(  k/n \right)^{2/d}$ and $1/(kT)$.  
% Balancing these terms under the budget constraint $nT=\Gamma$ yields  
% \[
% T^*\asymp \Gamma^{2/\left(d+2\right)}/k^*, 
% \qquad 
% n^*\asymp \Gamma^{d/\left(d+2\right)} k^*,
% \qquad 
% k^*T^* \asymp \Gamma^{2/\left(d+2\right)},
% \qquad
% (k^*)^{-1}=o(1).
% \]
% Since $T\gtrsim k$, we further require $T^* \gtrsim \Gamma^{1/\left(d+2\right)}$.  
% Thus the admissible order of $T^*$ lies in $\left.\left[\Gamma^{1\over d+2}, \Gamma^{2\over d+2}\right)\right.$. 
% Substituting into Equation~\eqref{eq:kNN_MSE} gives the optimal order of the componentwise MSE,  $\oE\left[\left(\hat{\theta}_{kNN}^j\left( x \right)-\theta^j_*\left( x \right)\right)^2\right]=\tilde{O}\left(\Gamma^{-{2\over d+2}}\right)$.

\vspace{6pt}

\emph{Case 2: $T=o(k)$.} 
The dominant terms are $\left(  k/n \right)^{2/d}$ and $1/T^2$, and since $\left(  k/n \right)^{2/d}+1/T^2$ increases with $k$, 
\[
\left(  k/n \right)^{2/d}+1/T^2>(T/n)^{2/d}+1/T^2\ge \Gamma^{-2/\left(d+2\right)},
\]
with equality only if $T=\Gamma^{1/\left(d+2\right)}$. 
Thus, the MSE rate here is strictly worse than in Case~1. Therefore, the optimal order of $\oE\left[\left(\hat{\theta}_{kNN}^j\left( x \right)-\theta^j_*\left( x \right)\right)^2\right]$ is $\tilde{O}\left(\Gamma^{-{2\over d+2}}\right)$, achieved in Case 1.

% \emph{Case 2: $T=o(k)$.}  
% Here the dominant terms are $\left(  k/n \right)^{2/d}$ and $1/T^2$.  
% Notice that $\left(  k/n \right)^{2/d}+1/T^2$ is increasing in $k$. 
% Thus,
% \[\left(  k/n \right)^{2/d}+1/T^2
% >(T/n)^{2/d}+1/T^2
% \geq \Gamma^{-{2\over d+2}},\]
% where the equality in the last inequality holds if and only if $T=\Gamma^{1\over d+2}$.
% Hence the MSE rate in this case is strictly worse than in Case 1.

By Proposition~\ref{prop:MSE_decom}, the aggregated MSE satisfies
$$\oE\left[\left\|\hat\theta_{kNN}\left( x \right)-\theta_*\!\left(x\right)\right\|^2\right]\asymp q \oE\left[\left(\hat{\theta}_{kNN}^j\left( x \right)-\theta^j_*\left( x \right)\right)^2\right] = \tilde{O}\left(q\Gamma^{-{2\over d+2}}\right).$$
Finally, by Corollary~\ref{col:convergence}, the optimality gap  $f\left(\hat{\theta}_{kNN}\left( x \right);x\right)- f\left(\theta\!\left(x\right);x\right)\to 0$ in probability and converges in the rate of $\tilde{O}_{\rm P}\left(q\Gamma^{-{2\over d+2}}\right)$.
\hfill\Halmos
\end{proof}

\vspace{6pt}

% Theorem~\ref{thm:kNN} yields three key insights: First, the allocation of the per-covariate simulation budget $T$ is crucial because it directly governs the solution accuracy of $\bar{\theta}_T\left(x_i\right)$. Specifically, the order of $T^*$ not only determines the sample size $n^*=\Gamma/T^*$ but also dictates the hyperparameter choice $k^*\asymp \Gamma^{2/\left(d+2\right)}/T^*$. 
% Actually, the order of optimal per-covariate simulation effort, $T^*$, lies in the interval $\left.\left[\Gamma^{1\over d+2}, \Gamma^{2\over d+2}\right)\right.$, rather than being fixed at a single order. The lower bound ensures sufficient suppression of the solution-induced bias from inexact optimization, while the upper bound is implicitly enforced by the need to maintain enough covariate coverage  (i.e., a sufficiently large $n$) to control the interpolation bias term $\left(  k/n \right)^{1/d}$.
% This highlights a fundamental trade-off: improving the accuracy of  $\bar{\theta}_T\left(x_i\right)$ through larger $T$ is ultimately constrained by the need for broad covariate coverage, $n$.

Theorem~\ref{thm:kNN} yields three key insights. 
First, the optimal range of the PR-SGD iteration number, $T^*$, provides insights into how ``exact'' the offline optimization should be to achieve the best convergence rate. 
On one hand, the lower bound of $T^*$ ensures that the solution bias, i.e., the bias introduced by using a finite number of PR-SGD iterations, is sufficiently small and does not dominate the overall estimation error.
On the other hand, the upper bound of $T^*$ guarantees that the number of designed covariates $n=\Gamma/T$ remains large enough to ensure adequate coverage of the covariate space, thereby controlling the interpolation bias of the kNN approximation.
This highlights a fundamental trade-off: enhancing the accuracy of the offline solutions  $\bar{\theta}_T\left(x_i\right)$'s through a larger $T$ is inevitably constrained by the need to maintain sufficient covariate coverage, $n$.

Second, both of the optimal convergence rate and the optimal sample-allocation rule depend on the dimensionality of covariate, $d$. The optimal convergence rate, $q\Gamma^{-{2\over d+2}}$, suffers from the curse of dimensionality. The optimal sample-allocation rule shows that allocating more budget to $n$ (for better covariate coverage) is typically more beneficial than $T$ (for more accurate local solutions) in the high-dimensional settings.

% Third, the choice of $k$. The optimal rule $k^*T^* \asymp \Gamma^{2/\left(d+2\right)}$ provides a novel interpretation of the estimator as a $kT$-nearest neighbor method, which aggregates information from $k$ neighboring covariates, each contributing an average over $T$ simulations. 

Third, different from conventional kNN technique that achieves the optimal convergence rate solely by tuning $k$, our algorithm follows the allocation rule $k^*T^* \asymp \Gamma^{2/\left(d+2\right)}$, which treats the product $kT$ as an integrated quantity. This yields a new perspective: the kNN estimator based on the $T$-step PR-SGD solutions can be viewed as an effective $kT$-nearest neighbor technique. The key insight is that the algorithm aggregates information from $k$ neighbors, each of which represents an averaged solution obtained from $T$ stochastic simulations. In this way, statistical and optimization errors are coupled.

Lastly, we want to make two additional remarks on the analysis framework that leads to Theorem~\ref{thm:kNN} and the importance of the $\tilde{O}\left(1/T\right)$ bias of the PR-SGD solution, established in Proposition~\ref{prop:bias}.

\begin{remark}
\textit{The analysis framework used to derive the convergence rate of optimality gap and the optimal sample-allocation rule in Theorem~\ref{thm:kNN} is applicable beyond the KNN estimator. It can be used to analyze any smoothing techniques employed in Algorithm~\ref{alg}. In the rest of this section, we apply the same framework to derive the optimal convergence rate of the optimality gap for other smoothing techniques. 
}
\end{remark}

\vspace{6pt}

\begin{remark}
\textit{If the solution bias is only treated as $o\left(1/\sqrt{T}\right)$, e.g., of order $T^{-1/2-\delta}$ for some unknown constant $\delta>0$, as implied by asymptotic normality, then, following the same analysis procedures as in the proof of Lemma~\ref{lem:kNN} and Theorem~\ref{thm:kNN}, the order of $T^*$ lies in the interval $\left[\Gamma^{\tfrac{2}{\left(d+2\right)(1+2\delta)}},\, \Gamma^{\tfrac{2}{d+2}}\right)$.  
However, since the exact value of $\delta$ is unknown, the precise order of $T^*$ cannot be determined.  
This ambiguity highlights the necessity of deriving a sharper bound on the bias, as carried out in Proposition~\ref{prop:bias}.}
\end{remark}

\subsection{Kernel Smoothing}
KS estimates the target function at $x$  by a weighted average of the observed responses in the neighborhood of $x$, where the weights are determined by a kernel function of the distance between $x$ and each sample point $x_i$.
A typical kernel estimator, the Nadaraya-Watson estimator ~\citep{nadaraya1964estimating,watson1964smooth} is given by:
\begin{equation}\label{eq:KS}
\hat\theta_{KS}\left( x \right)
=\frac{\sum_{i=1}^n k_h\left(x-x_i\right)\bar{\theta}_T\left(x_i\right)}{\sum_{i=1}^nk_h\left(x-x_i\right)},
% =\sum_{i=1}^n w\!\left(x_i,\, x\right)\bar{\theta}_T\left(x_i\right)
% =\sum_{i=1}^n w_i\left( x \right)\left(\theta^*\left( x \right) + {\xi\left( x \right)\over\sqrt{T}} + {\varepsilon\left( x \right)\over T}\right),    
\end{equation}
where $k_h\left( x \right)=\left( 1/h^d\right)K\left( x/h \right)$, $K$ is a symmetric density function and $h$ is the bandwidth parameter satisfying $h\to 0$ and $nh^d\to \infty$ as $n\to\infty$. 
Hence, $w\!\left(x_i,\, x\right) =k_h\left(x-x_i\right)/\sum_{i=1}^n k_h\left(x-x_i\right)$ in the KS estimator.
We adopt a sphere kernel, $K\left( x,x_i\right)=I\{\|x-x_i\|\leq 1\}$, in the analysis, as it is known that the choice of kernel function $K$ is not critical in estimation \citep{hong2017kernel}.
We summarize the MSE of $\hat{\theta}^j_{KS}\left( x \right)$ in Lemma~\ref{lem:KS} and provide the detailed proof in Appendix~\ref{app:KS_lem}, which follows a similar proof sketch as Lemma~\ref{lem:kNN}.

\begin{lemma}[KS]\label{lem:KS}
Suppose Assumptions~\ref{as:smoothness}–\ref{as:3.4} hold.
Consider Algorithm~\ref{alg}, in which the offline SO problems are solved via a $T$-step PR-SGD algorithm and the optimal-solution function is approximated by KS based on the sample $\left\{x_i, \bar{\theta}_T\left(x_i\right)\right\}_{i=1}^n$.
% , where $\bar{\theta}_T\left(x_i\right)$'s are the offline solutions at the designed covariates.
Then, for every dimension $j=1,\ldots,q$, the MSE of predicted solution satisfies  
\[
\oE\left[\left(\hat{\theta}_{KS}^j\left( x \right)-\theta^j_*\left( x \right)\right)^2\right] 
= \tilde{O}\left(h^2+{1\over T^2}+{1\over nT h^d}\right),  \ \text{ as }\ n \text{ and } T\to \infty.
\]
\end{lemma}

The KS estimator with the sphere kernel utilizes the solutions $\bar{\theta}_T\left(x_i\right)$'s such that $\|x_i-x\|\leq h$. Its variance is determined by the number of $x_i$'s falling within this  bandwidth-defined neighborhood.
Its bias arises from two sources: the solution bias from the $T$-step PR-SGD algorithm and the interpolation bias depending on the bandwidth $h$. To determine the optimal sample-allocation rule and the choice of $h$,  we solve the following optimization problem and summarize the results in Theorem~\ref{thm:KS}. The detailed proof is included in Appendix~\ref{app:KS_thm}.
\begin{eqnarray*}
\min_{n,T,h}\ h^2+{1\over T^2}+{1\over nTh^d}, \quad
 \text{subject to}\quad  nT=\Gamma,\ h=o(1),\ \left(nh^d\right)^{-1}=o(1).
\end{eqnarray*}

\begin{theorem}[KS]
    \label{thm:KS}
Suppose Assumptions \ref{as:smoothness}-\ref{as:3.4} hold.
Consider Algorithm~\ref{alg}, in which the offline SO problems are solved via a $T$-step PR-SGD algorithm and the optimal-solution function is approximated by KS. 
Giving a total simulation budget $\Gamma$, the optimal order of hyperparameter is $h^*\asymp\Gamma^{-{1\over d+2}}$. The order of $T^*$ lies in the interval $\left.\left[\Gamma^{1\over d+2}, \Gamma^{2\over d+2}\right)\right.$ and the order of $n^*=\Gamma/T^*$ lies in the interval $\left.\left(\Gamma^{d\over d+2}, \Gamma^{d+1\over d+2}\right]\right.$.
Under this sample-allocation rule and hyperparameter setting, the optimality gap $f\left(\hat{\theta}_{KS}\left( x \right);x\right)- f\left(\theta\!\left(x\right);x\right)\to 0$ in probability and converges in the rate of $\tilde{O}_{\rm P}\left(q\Gamma^{-{2\over d+2}}\right)$.
 % the optimal convergence rate  of the optimality gap is $q\tilde O_{\rm P}\left(\Gamma^{-{2\over d+2}}\right)$.
\end{theorem}

We offer several remarks on the Theorem~\ref{thm:KS}.
First, similar to Theorem~\ref{thm:kNN}, the optimal sample-allocation rule requires a lower bound on iteration number of PR-SGD algorithm, $T$, to suppress the impact of solution bias. However, excessive exactness, specifically, choosing $T \gtrsim \Gamma^{2/\left(d+2\right)}$ is suboptimal, as it compromises the coverage of the covariate space.
% , which requires that  the order of $n^*$ is strictly larger than $\Gamma^{d\over d+2}$.
Again, the convergence rate $q\Gamma^{-{2\over d+2}}$ suffers from the curse of dimensionality. In high-dimensional settings, expanding covariate coverage by increasing $n$ is more beneficial than improving per-covariate exactness by using a larger number of iteration steps, $T$, in the PR-SGD algorithm.

Second, analogous to the role of $k$ in the kNN technique, the bandwidth parameter $h$ plays a central role in determining the estimator's performance. A smaller $h$ produces more localized but higher-variance estimators while a larger one induces smoother estimators with higher bias. The optimal bandwidth satisfies $h^*\asymp\Gamma^{-{1\over d+2}}$.

Third, it is not surprising that the kNN and KS estimators exhibit the same convergence rate and requirements on $T$. This is because kNN can be interpreted as a kernel smoother with adaptive bandwidth \citep{hastie2009elements}. More details about their connection can be found in  Remark~\ref{remark} in the Appendix ~\ref{app:smoothing}.

\subsection{Linear Regression}
LR is a parametric technique, aiming to find the best-fitting linear model that describes the relationship between a dependent variable and the basis functions of independent variables. 
Consider a set of basis functions \(\phi\left( x \right) = (\phi_1\left( x \right), \dots, \phi_s\left( x \right))^\top \in \mathbb{R}^s\). Suppose that the true optimal-solution function can be modeled as $\theta_*\!\left(x\right)=\phi\left( x \right)^\top\beta+M\left( x \right)$, where  $\beta^\top \phi\left( x \right)$ is the best linear approximation afforded by the basis functions $\phi\left( x \right)$, $M\left( x \right)$ is the residual error under such approximation and $\beta =(\beta^1,\ldots,\beta^q) \in \mathbb{R}^{s\times q}$ with $\beta^j\in\mathbb{R}^{s}$ is the coefficient matrix. 
Although the irreducible bias $M\left( x \right)$ does not vanish with increased computation effort, from a practical point of view,  it can typically be kept small by selecting appropriate basis functions, especially when the problem dimension is low. Therefore, we treat it as $0$ here.

The coefficient $\beta$ is estimated by solving the least squares problem 
$\min_{\beta} \frac{1}{n} \sum_{i=1}^n \left\|\bar\theta_T^j\left(x_i\right) - \phi\left(x_i\right)^\top\beta ^j\right\|^2$
and can be represented by $\hat{\beta}=\left(\Phi^\top \Phi\right)^{-1}\Phi^\top \bar{\theta}_T$ with $\Phi = \left(\phi\left(x_1\right), \ldots, \phi\left( x_n \right)\right)^\top\in\mathbb{R}^{n\times s}$ and $\bar{\theta}_T = \left(\bar{\theta}_T\left(x_1\right),\ldots,\bar{\theta}_T\left( x_n \right)\right)^\top\in \mathbb{R}^{n\times q}$ with $\bar{\theta}_T\left(x_i\right)=\left(\bar{\theta}^1_T\left(x_i\right),\ldots,\bar{\theta}^q_T\left(x_i\right)\right)^\top \in \mathbb{R}^{q}$. 
Then, the LR estimator at the target point $x$ is given by 
\begin{equation}\label{eq:LR}
\hat\theta_{LR}\left( x \right) =  \hat \beta^\top \phi\left( x \right) 
= \bar{\theta}_T^\top \Phi  \left(\Phi^\top \Phi\right)^{-1} \phi\left( x \right)
=w\left( x \right)^\top \bar{\theta}_T
=\sum_{i=1}^n w\!\left(x_i,\, x\right)\bar{\theta}_T\left(x_i\right),  
\end{equation}
by letting $w\left( x \right)=\Phi\left(\Phi^\top\Phi\right)^{-1}\phi\left( x \right)\in \mathbb{R}^n$ and the weight $w\!\left(x_i,\, x\right)$ be the $i$-th entry of the vector $w\left( x \right)$. 
We summarize the MSE of $\hat{\theta}^j_{LR}\left( x \right)$ in Lemma~\ref{lem:LR}. The proof is deferred to Appendix~\ref{app:LR_lem}.

\begin{lemma}[LR]\label{lem:LR}
Suppose Assumptions~\ref{as:smoothness}–\ref{as:3.4} hold.
Consider Algorithm~\ref{alg}, in which the offline SO problems are solved via a $T$-step PR-SGD algorithm and the optimal-solution function is approximated by LR based on the sample $\left\{x_i, \bar{\theta}_T\left(x_i\right)\right\}_{i=1}^n$.
Then, for every dimension $j=1,\ldots,q$, the MSE of predicted solution satisfies 
\[
\oE\left[\left(\hat{\theta}_{LR}^j\left( x \right)-\theta^j_*\left( x \right)\right)^2\right] 
= \tilde{O}\left({1\over T^2}+{1\over nT}\right),  \ \text{ as }\ n \text{ and } T\to \infty.
\]
\end{lemma}

Unlike the kNN and KS estimators, the bias of the LR estimator arises solely from the solution bias, as the linear regression is inherently an unbiased parametric technique, i.e., $\oE\left[\hat{\beta}^j\right]=\beta^j$. 
Analogous to classical linear regression, the variance is determined by the sample size $n$, with each contributing an independent variance of order $1/T$.
To determine the optimal sample-allocation rule,  we solve the following optimization problem and summarize the results in Theorem~\ref{thm:LR}. The detailed proof is included to Appendix~\ref{app:LR_thm}.
\begin{eqnarray*}
\min_{n,T} {1\over T^2}+{1\over nT}, \quad
 \text{subject to}\quad  nT=\Gamma.
\end{eqnarray*}

\begin{theorem}[LR]
    \label{thm:LR}    
Suppose Assumptions \ref{as:smoothness}-\ref{as:3.4} hold.
Consider Algorithm~\ref{alg}, in which the offline SO problems are solved via a $T$-step PR-SGD algorithm and the optimal-solution function is approximated by LR. 
Giving  a total simulation budget $\Gamma$, the optimal sample-allocation rule satisfies 
$T^*\gtrsim \Gamma^{1\over 2}$ and $n^*=\Gamma/T^*$.
Under this sample-allocation rule and hyperparameter setting, the optimality gap $f\left(\hat{\theta}_{LR}\left( x \right);x\right)- f\left(\theta\!\left(x\right);x\right)\to 0$ in probability and converges in the rate of $\tilde{O}_{\rm P}\left(q\Gamma^{-1}\right)$.
\end{theorem}

We offer several remarks on Theorem~\ref{thm:LR}.
First, the optimal sample-allocation rule requires a lower bound on the iteration number of PR-SGD algorithm, $T \gtrsim \sqrt{\Gamma}$, to control the solution bias.
Unlike kNN or KS, there is no trade-off between multiple sources of bias: the LR estimator incurs only solution-induced bias, and thus does not impose  upper bound on $T^*$  or lower bound on $n^*$ to control the interpolation bias. Practically, the only structural requirement is  \(n \geq s\) to ensure identifiability in linear regression \citep{hastie2009elements}.  

Second, the optimal convergence rate is of order $q\Gamma^{-1}$, which is free from the impact of dimensionality and is the fastest rate attainable by any classical regression technique. This rate is dictated solely by the variance term of order ${1\over nT}$ with $nT=\Gamma$, provided that the solution bias is adequately controlled. However, we emphasize that the irreducible bias $M(x)$ is assumed to be zero here through an appropriate selection of basis functions. Otherwise, the optimality gap would fail to converge to zero.

\subsection{Kernel Ridge Regression}\label{subsec:krr}
KRR is a non-parametric technique that combines ridge regression with the kernel trick, enabling it to effectively capture non-linear relationships. 
It involves solving the optimization problem 
\begin{equation}\label{eq:KRR_opt}
\min_{\hat{\theta}^j\in \mathcal{H}_K} {1\over n}\sum_{i=1}^n \left(\bar{\theta}_T^j\left(x_i\right)-\hat{\theta}^j\left(x_i\right)\right)^2 + \lambda\left\|\hat{\theta}^j\right\|^2_{\mathcal{H}_K}, \ j=1,\ldots,q,
\end{equation}
where $\lambda>0$ is the regularization parameter and
$\mathcal{H}_K$ is the reproducing kernel Hilbert space (RKHS)  induced by the positive semidefinite kernel $K$. Correspondingly, $\|\cdot\|_{\mathcal{H}_K}$ is the norm defined in the RKHS $\mathcal{H}_K$, which reflects the overall complexity of the function. 
Note that different from classical ridge regression, which estimates model parameters, Problem~\eqref{eq:KRR_opt} is formulated and solved in the function space $\mathcal{H}_K$ to directly obtain the estimated function $\hat\theta^j$. 
By the representation theorem \citep{scholkopf2001generalized}, the optimal solution of Problem~\eqref{eq:KRR_opt}, i.e., the KRR estimator, is given by 
\begin{equation}\label{eq:KRR}
\hat\theta_{KRR}\left( x \right)= r\left( x \right)^\top \left(R+n\lambda I\right)^{-1}\bar{\theta}_T
=w\left( x \right)^\top \bar{\theta}_T
=\sum_{i=1}^n w\!\left(x_i,\, x\right) \bar{\theta}_T\left(x_i\right),  
\end{equation}
where $r\left( x \right) = \left(K\left( x,x_1\right),\ldots, K\left( x,x_n\right)\right)\in \mathbb{R}^{n}$, $R =  \left(K\left(x_i,x_j\right)\right)_{i,j=1}^n \in \mathbb{R}^{n\times n}$ and $\bar{\theta}_T = (\bar{\theta}_T\left(x_1\right),\ldots,\bar{\theta}_T\left( x_n \right))^\top\in \mathbb{R}^{n\times q}$ with $\bar{\theta}_T\left(x_i\right)=\left(\bar{\theta}^1_T\left(x_i\right),\ldots,\bar{\theta}^q_T\left(x_i\right)\right)^\top \in \mathbb{R}^{q}$, .
The weight vector is  $w\left( x \right)=\left(R+n\lambda I\right)^{-1} r\left( x \right)\in \mathbb{R}^n$ and the weight $w\!\left(x_i,\, x\right)$ is the $i$-th entry of the vector $w\left( x \right)$.

According to Proposition~\ref{prop:smoothness}, i.e., $\theta_*\!\left(x\right)\in C^{m}$, and the compactness of  $\mathcal{X}$, it follows that $\theta_*\!\left(x\right)$ belongs to the Sobolev space of order $m$.
Since the RKHS associated with the Mat\'ern kernel of smoothness $\nu$ is norm-equivalent to the Sobolev space of order $\nu+d/2$ \citep{kanagawa2018gaussian}, we adopt the Mat\'ern kernel with $\nu=m-d/2$ to match the smoothness of $\theta_*\!\left(x\right)$, which is the standard choice under an additional smoothness assumption $m>d/2$ when KRR is applied. 
We summarize the MSE of $\hat{\theta}^j_{KRR}\left( x \right)$ in Lemma~\ref{lem:KRR}. 
% The detailed proof is deferred to Appendix~\ref{app:KRR_lem}.

\begin{lemma}[KRR]\label{lem:KRR}
Suppose Assumptions~\ref{as:smoothness}–\ref{as:3.4} hold.
Consider Algorithm~\ref{alg}, in which the offline SO problems are solved via a $T$-step PR-SGD algorithm and the optimal-solution function is approximated by KRR based on the sample $\left\{x_i, \bar{\theta}_T\left(x_i\right)\right\}_{i=1}^n$.
Then, for every dimension $j=1,\ldots,q$, the MSE of predicted solution satisfies  
\[
\oE\left[\left(\hat{\theta}_{KRR}^j\left( x \right)-\theta^j_*\left( x \right)\right)^2\right] 
= \tilde{O}\left(\lambda^{1-{d\over 2m}}+T^{-2}+n^{-1}T^{-1}\lambda^{-{d\over2m}}\right), \ \text{ as }\ n \text{ and } T\to \infty.
\]
\end{lemma}
\begin{proof}{Proof.}
Based on Equation~\eqref{eq:KRR}, the bias decomposes into the following  two parts:
\[
\oE\left[\theta_{KRR}^j\left( x \right)\right]-\theta_*^j\left( x \right)
=w\left( x \right)^\top\oE\left[\bar{\theta}_T^j\right]-\theta_*^j\left( x \right)
=\underbrace{w\left( x \right)^\top\left(\oE\left[\bar{\theta}_T^j\right]-\theta_*^j\right)}_{{\text{solution-induced bias}}}
+\underbrace{\left(w\left( x \right)^\top \theta_*^j-\theta_*^j\left( x \right)\right)}_{\text{interpolation bias}},
\]
where 
% $w\left( x \right) = \left(R+n\lambda I\right)^{-1}r\left( x \right)\in \mathbb{R}^n$,
$\bar{\theta}_T^j = \left(\bar{\theta}^j_T\left(x_1\right), \ldots, \bar{\theta}^j_T\left( x_n \right)\right)^\top$ and $\theta^j_* = \left(\theta^j_*\left(x_1\right), \ldots, \theta^j_*\left( x_n \right)\right)^\top$.

By Proposition~\ref{prop:bias}, the solution-induced bias satisfies $w\left( x \right)^\top\left(\oE\left[\bar{\theta}_T^j\right]-\theta_*^j\right)
=\tilde{O}\left({1\over T}\right)w\left( x \right)^\top \bm 1_n$.
We first derive the order of $s\left( x \right) = w\left( x \right)^\top \bm 1_n=r\left( x \right)^\top\left(R+n\lambda I\right)^{-1}\bm{1}_n$.
Consider the function approximation problem with target function \(g\equiv 1\) and the dataset $\{(x_i, 1)\}_{i=1}^n$, where $\{x_i\}_{i=1}^n$ are quasi-uniform.
Denote the approximated function provided by KRR as $\hat g$ and the point estimator at $x$ is $\hat g\left( x \right)=r\left( x \right)^\top\left(R+n\lambda I\right)^{-1}\bm{1}_n$.
By Theorem 7 of \citet{tuo2024asymptotic}, $|\hat g\left( x \right)-g\left( x \right)|=|\hat g\left( x \right)-1|\lesssim \lambda^{{1\over 2}-{d\over 4m}}$ if $\lambda=o(1)$ and $\lambda^{-1}=o\left(n^{2m\over d}\right)$.

Notice that the representation of $\hat g\left( x \right)=r\left( x \right)^\top\left(R+n\lambda I\right)^{-1}\bm{1}_n$ coincides with $s\left( x \right) = w\left( x \right)^\top \bm 1_n$.
Then, $s\left( x \right)=1+s\left( x \right)-1=1+\hat{g}\left( x \right)-g\left( x \right)\lesssim 1+\lambda^{{1\over 2}-{d\over 4m}}$, which implies $s\left( x \right)=w\left( x \right)^\top \bm 1_n = 1+o(1)$. Therefore, the solution-induced bias satisfies
\[
w\left( x \right)^\top\left(\oE\left[\bar{\theta}_T^j\right]-\theta_*^j\right) = \tilde{O}\left({1\over T}\right)s\left( x \right)
= \tilde{O}\left({1\over T}\right)(1+o(1)) = \tilde{O}\left({1\over T}\right).
\]
The interpolation bias is bounded by Theorem 7 of \citet{tuo2024asymptotic} as
\[
w\left( x \right)^\top \theta_*^j-\theta_*^j\left( x \right)
= r\left( x \right)^\top\left(R+n\lambda I\right)^{-1}\theta^j_*-\theta^j_*\left( x \right)
\lesssim\lambda^{{1\over2}-{d\over 4m}}.
\]
Thus, the overall bias $\oE\left[\theta_{KRR}^j\left( x \right)\right]-\theta_*^j\left( x \right) = \tilde{O}\left({1\over T}+\lambda^{{1\over2}-{d\over 4m}}\right)$.

For the variance,
\[
\oV\left[\theta_{KRR}^j\left( x \right)\right]
= \oV\left[w\left( x \right)^\top\bar{\theta}_T^j\right]
=w\left( x \right)^\top \oV\left[\bar{\theta}_T^j\right]w\left( x \right)
\asymp {1\over T}w\left( x \right)^\top I w\left( x \right).
\]
Notice that  $w\left( x \right)^\top I w\left( x \right) = r\left( x \right)^\top\left(R+n\lambda I\right)^{-2}r\left( x \right)\asymp n^{-1}\lambda^{-{d\over2m}}$ according to Theorem 7 of \citet{tuo2024asymptotic}. Then, the variance
$\oV\left[\theta_{KRR}^j\left( x \right)\right]\asymp {1\over nT}\lambda^{-{d\over2m}}$.

Combining the bias and variance yields, $$\oE\left[\left(\hat{\theta}_{KRR}^j\left( x \right)-\theta^j_*\left( x \right)\right)^2\right] 
= \tilde{O}\left({1\over T^2}+\lambda^{1-{d\over 2m}}\right) + O\left({1\over nT}\lambda^{-{d\over2m}}\right)
= \tilde{O}\left({1\over T^2}+\lambda^{1-{d\over 2m}}+{1\over nT}\lambda^{-{d\over2m}}\right),$$
for all $j=1,\ldots,q$, which gives the stated result.
\hfill\Halmos
\end{proof}

\smallskip

Analogous to the kNN and KS estimators, the overall bias for KRR estimator has two components: an interpolation bias, governed by the regularization parameter $\lambda$, 
% $\lambda^{{1\over 2}-{d\over 4m}}$ 
and a solution-induced bias, governed by the per–covariate effort $T$.
% , which is of order $T^{-1} \lambda^{-{d\over4m}}$.
A key distinction is that, in KRR, the weights generally do not sum to one, i.e., $\sum_{i=1}^n w\!\left(x_i,\, x\right)\neq 1$.
Nevertheless, with an appropriate choice of $\lambda$, we prove that $\sum_{i=1}^n w\!\left(x_i,\, x\right)=1+o(1)$. Consequently, the solution-induced bias remains of the same order as the offline solution bias, $\tilde{O}\left(1/T\right).$
The variance is governed by the total computation effort $nT$ and the regularization parameter $\lambda$.
Structurally, this is highly similar to KS, where the variance is controlled by $nT$ and the the bandwidth parameter $h$ while the interpolation bias is determined by $h$. Indeed, KRR can be interpreted as a kernel smoother with an equivalent kernel, in which the effective bandwidth is controlled by $\lambda$ \citep{williams2006gaussian}.
For a fixed computation budget $\Gamma$, the optimal convergence rate of the MSE can be obtained by solving the following optimization problem  and the results are summarized in Theorem~\ref{thm:KRR}.
\begin{eqnarray*}
\min_{n,T,\lambda} \ {1\over T^2}+\lambda^{1-{d\over 2m}}+{1\over nT}\lambda^{-{d\over2m}}, \quad
 \text{subject to}\quad nT=\Gamma,\ \lambda=o(1),\ \lambda^{-1}=o\left(n^{2m\over d}\right).
\end{eqnarray*}

\begin{theorem}[KRR]
    \label{thm:KRR}
Suppose Assumptions \ref{as:smoothness}-\ref{as:3.4} hold.
Consider Algorithm~\ref{alg}, in which the offline SO problems are solved via a $T$-step PR-SGD algorithm and the optimal-solution function is approximated by KRR. 
Giving  a total simulation budget $\Gamma$, and under Assumptions \ref{as:smoothness}-\ref{as:3.4}
the optimal order of hyperparameter is $\lambda \asymp \Gamma^{-1}$. The order of $T^*$ lies in the interval $\left.\left[\Gamma^{{1\over 2}-{d\over 4m}}, \Gamma^{1-{d\over 2m}}\right)\right.$ and the order of $n^*=\Gamma/T^*$ lies in the interval $\left.\left(\Gamma^{{d\over 2m}},\Gamma^{{1\over 2}+{d\over 4m}}\right]\right.$.
Under this sample-allocation rule and hyperparameter setting, the optimality gap $f\left(\hat{\theta}_{KRR}\left( x \right);x\right)- f\left(\theta\!\left(x\right);x\right)\to 0$ in probability and converges in the rate of $\tilde O_{\rm P}\left(q\Gamma^{-{2m-d\over 2m}}\right)$.
\end{theorem}

\begin{proof}{Proof.}
Notice that the squared interpolation bias $\lambda^{1-{d\over 2m}}$ and the variance ${1\over \Gamma}\lambda^{-{d\over2m}}$ do not depend on the allocation scheme of $n$ and $T$.
Balancing these two terms yields $\lambda^*\asymp\Gamma^{-1}$ and 
the corresponding componentwise MSE of order $\tilde{O}\left(\Gamma^{-{2m-d\over 2m}}\right)$.
% $\left(\lambda^*\right)^{1-{d\over 2m}}+ {1\over \Gamma}\left(\lambda^*\right)^{-{d\over2m}}\asymp\Gamma^{-{2m-d\over 2m}}$
To ensure that the overall MSE is not dominated by the solution-induced bias, it also requires that $T^2\gtrsim \Gamma^{{2m-d\over 2m}}$, i.e., $T^*\gtrsim\Gamma^{{2m-d\over 4m}}$.
Meanwhile, the constraint $\lambda^{-1}=o\left(n^{2m\over d}\right)$ requires  $n^{-1}=o\left(\Gamma^{-{d\over 2m}}\right)$.
Therefore, the order of $T^*$ lies in the interval $\left.\left[\Gamma^{{1\over 2}-{d\over 4m}}, \Gamma^{1-{d\over 2m}}\right)\right.$ and $n^*$ lies in the interval $\left.\left(\Gamma^{{d\over 2m}},\Gamma^{{1\over 2}+{d\over 4m}}\right]\right.$ with the resulting optimal componentwise MSE of order $\Gamma^{-{2m-d\over 2m}}$.

Similarly, by Proposition~\ref{prop:MSE_decom} and Corollary~\ref{col:convergence}, the optimality gap converges in the rate of $\tilde O_{\rm P}\left(q\Gamma^{-{2m-d\over 2m}}\right)$.
\hfill \Halmos
\end{proof}

\smallskip

We provide several remarks on Theorem~\ref{thm:KRR}.
First, regarding sample allocation, KRR shares a similar structure with kNN and KS, where the optimal sample-allocation rule imposes both lower and upper bounds on the iteration number of PR-SGD algorithm, $T$, with the lower bound decreasing as the dimension \(d\) increases.  
The key difference lies in the role of smoothness \(m\).  
When \(m\) is large, i.e., the optimal-solution function is sufficiently smooth, the requirement on coverage \(n^*\) is relaxed, allowing more budget to be allocated to solve each SO problem, i.e., larger $T^*$.  
In the extreme case of sufficiently large \(m\), the requirement on \(T^*\) approaches that of LR, namely \(T^* \gtrsim \sqrt{\Gamma}\).  Second, in terms of convergence rate, KRR is affected by dimensionality \(d\) as in kNN and KS, but it also explicitly benefits from smoothness \(m\).  
Once \(m > d/2 + 1\), the convergence rate of KRR surpasses that of kNN and KS.  
As \(m\) grows to infinity, the convergence rate approaches \(q\Gamma^{-1}\), matching the performance of LR and effectively avoiding the curse of dimensionality.

\subsection{Discussions}\label{subsec:discussion}
To conclude this section, we summarize and discuss the theoretical properties of the four smoothing techniques, highlighting their implications for both theory and practice.

\subsubsection{Impact of Solution Bias.}
The smoothing techniques analyzed in this section are used in classical regression (CR) settings as well. A key difference between our setting and CR lies in the bias of the training data. In our problem, the solutions produced by the $T$-step PR-SGD algorithm are biased, with variances on the order of $1/T$. In contrast, in CR where $T$ observations can be taken at the same design point, the sample means are unbiased with variances of the same order.

To clarify the role of this bias, in Appendix~\ref{app:smoothing} we derive the optimal allocations and the optimal convergence rates for the unbiased setting (denoted CR). These results are compared with those obtained in Sections~\ref{subsec:knn} to \ref{subsec:krr} for the biased setting (denoted OTP) in Table~\ref{table:allocation}. Two observations emerge.
First, when bias is present (OTP), there is less flexibility in the choice of $T^*$. Although the upper bounds of $T^*$ appear same across the two settings, the OTP problem requires a larger lower bound to control bias, a pattern consistent across all four smoothing techniques. Second, despite the presence of bias in OTP, the optimal convergence rates coincide with those in the unbiased CR setting. This suggests that, under optimal sample-allocation rule, the bias inherent in PR-SGD solutions does not substantially degrade prediction quality.

\begin{table}[th]
\caption{Optimal sample allocations and optimal convergence rates of CR and OTP}\label{table:allocation}
\centering
\begin{tabular}{lllll}
\toprule
    & \multicolumn{2}{c}{Optimal Sample  Allocation}     & \multicolumn{2}{c}{Convergence Rate}\\ \cmidrule{2-3}  \cmidrule(l){4-5}
\multicolumn{1}{c}{Technique}     & \multicolumn{1}{c}{CR}      & \multicolumn{1}{c}{OTP}        & \multicolumn{1}{c}{CR} & \multicolumn{1}{c}{OTP} \\ \cmidrule{1-1}\cmidrule(l){2-2}  \cmidrule(l){3-3} \cmidrule(l){4-4} \cmidrule(l){5-5}
kNN & \begin{tabular}[c]{@{}l@{}}$k^*T^*=\Gamma^{2\over d+2}$,  $k^*\in\left(1,\Gamma^{{2\over d+2}}\right]$,\\ $T^*\in\left[1,\Gamma^{{2\over d+2}}\right)$\end{tabular} & \begin{tabular}[c]{@{}l@{}}$k^*T^*=\Gamma^{2\over d+2}$,  $k^*\in\left(1,\Gamma^{{1\over d+2}}\right]$,\\ $T^*\in\left[\Gamma^{{1\over d+2}},\Gamma^{{2\over d+2}}\right)$\end{tabular} & $q\Gamma^{-{2\over d+2}}$             & $q\Gamma^{-{2\over d+2}}$           \\ \cmidrule{1-1}\cmidrule(l){2-5}
KS  & \begin{tabular}[c]{@{}l@{}}$h^* = \Gamma^{-{1\over d+2}}$,  \\ $T^*\in\left[1,\Gamma^{{2\over d+2}}\right)$ \end{tabular}                           & \begin{tabular}[c]{@{}l@{}}$h^* = \Gamma^{-{1\over d+2}}$, \\ $T^*\in\left[\Gamma^{{1\over d+2}},\Gamma^{{2\over d+2}}\right)$\end{tabular} & $q\Gamma^{-{2\over d+2}}$ & $q\Gamma^{-{2\over d+2}}$ \\ \cmidrule{1-1}\cmidrule(l){2-5}
LR  & $T^*\in[1,T]$ & $T^*\in\left[\Gamma^{1\over 2},\Gamma\right]$ & $q\Gamma^{-1}$   & $q\Gamma^{-1}$\\ \cmidrule{1-1}\cmidrule(l){2-5}
KRR & \begin{tabular}[c]{@{}l@{}}$\lambda^*=\Gamma^{-1}$, \\ $ T^*\in\left[1,\Gamma^{1-{d\over2m}}\right)$\end{tabular}            & \begin{tabular}[c]{@{}l@{}}$\lambda^*=\Gamma ^{-1}$, \\ $T^*\in\left[\Gamma^{{1\over2}-{d\over4m}},\Gamma^{1-{d\over2m}}\right)$\end{tabular}        & $q\Gamma^{-\left(1-{d\over2m}\right)}$       & $q\Gamma^{-\left(1-{d\over2m}\right)}$     \\ \bottomrule
\end{tabular}

\end{table}

\subsubsection{Quality of Offline and Online Solutions.} 
When exact solutions are available in the offline stage, the predicted online solutions are typically less accurate, as smoothing introduces interpolation error \citep{luo2024reliable}. One might expect this relationship to persist for inexact offline solutions, since the online stage learns from them. However, the results in Sections~\ref{subsec:knn} to \ref{subsec:krr} show a counterintuitive phenomenon: the online solutions are at least as accurate as the offline solutions, and sometimes even better.

Specifically, when the $T$-step PR-SGD algorithm is used to solve the offline problem at a given design point, Proposition~\ref{prop:bias} and Corollary~\ref{col:convergence} imply that the resulting optimality gap decreases at the rate $qT^{-1}$. Under the optimal allocation rules summarized in Table~\ref{table:allocation}, the predicted (online) optimality gaps converge at a rate that is no worse, and in some cases faster, than $qT^{*-1}$. Thus, the online solutions are guaranteed to be at least as good as the offline solutions.

This improvement arises because smoothing reduces part of the offline inexactness (specifically, the variance component), and this reduction more than compensates for the interpolation error introduced during smoothing. This mechanism is precisely the implication of Proposition~\ref{prop:MSE_decom}.

\subsubsection{Selection of Smoothing Techniques.}
Each of the four smoothing techniques exhibits its own advantages and trade-offs. kNN and KS are computationally simple and easy to implement, but their performance deteriorates in high-dimensional settings due to the curse of dimensionality. LR achieves the fastest convergence rate and is independent of the problem dimension, but its performance relies critically on the availability of suitable basis functions. When the structure of the optimal solution is complex or unknown, appropriate basis functions may be difficult to specify, and such misspecification introduces irreducible bias. KRR is capable of capturing complex functional structures and can mitigate the curse of dimensionality by exploiting higher-order smoothness, though it incurs higher computational costs due to matrix inversion and requires careful hyperparameter tuning.

In practice, the choice of smoothing technique is problem-dependent. For low-dimensional problems with ample computational budget, kNN and KS are often sufficient and easy to deploy. When reliable structural information allows for high-quality basis functions, LR is straightforward to apply and typically performs well. In other scenarios, KRR is generally recommended, as it can take advantage of the high-order smoothness commonly present in optimal-solution functions and requires little prior structural knowledge. Its main limitations lie in the need for hyperparameter tuning and the potentially significant computational burden when the number of design points is large.

\section{Numerical Study}\label{sec:num}

In this section, we use a test problem with practical features to validate the theoretical results of the proposed algorithmic framework, and to evaluate its practical performance. Following the theoretical analysis, we implement the four smoothing techniques, kNN, KS, LR, and KRR, and compare their performance using the empirical relative optimality gap. All experiments are implemented in Python and executed on a system equipped with two Intel Xeon Gold 6248R CPUs (each with 24 cores) and 256GB of RAM. The source code and complete experimental results are available in the GitHub repository available at \url{https://github.com/NiffyLin/CSCSO}.

Consider the following multi-product newsvendor problem:
\[
\min_{\theta_1,\ldots,\theta_q\geq 0} 
\mathbb{E}\!\left[
\sum_{i=1}^q \Big( 
c_{s_i}(D_i-\theta_i)^+ + 
c_{o_i}(\theta_i-D_i)^+
\Big)
\right],
\]
where product $i$ has order quantity $\theta_i$, demand $D_i$, shortage cost $c_{s_i}$ and overage cost $c_{o_i}$, for $i=1,\ldots,q$. We assume that the demands follow a linear factor model $D_i=\sum_{j=1}^d w_{ij} Z_j + \varepsilon_i$, $i=1,\ldots,q$, where $Z_1,\ldots,Z_d$ are common risk factors, $\varepsilon_1,\ldots,\varepsilon_q$ are idiosyncratic factors, and all factors are mutually independent. We further assume that $Z_j \sim N\left(x_j,\left(\gamma x_j\right)^2\right)$ and $\varepsilon_i \sim N\left(\mu_i,\left(\gamma \mu_i\right)^2\right)$ for $j=1,\ldots,d$ and $i=1,\ldots,q$.  Consequently, $D_i \sim N\left(
\sum_{j=1}^d w_{ij}x_j + \mu_i,\ 
\gamma^2\!\left(\sum_{j=1}^d w_{ij}^2 x_j^2 + \mu_i^2\right)
\right)$ for all $i=1,\ldots,q$. It then follows that the optimal solution is 
\[
\theta_i^* 
= \sum_{j=1}^d w_{ij} x_j + \mu_i 
+ z_{\alpha_i}\,
\gamma\sqrt{
\sum_{j=1}^d w_{ij}^2 x_j^2 + \mu_i^2
}, \quad i=1,\ldots,q,
\]
where $z_{\alpha_i}$ is the $\alpha_i$-quantile of the standard normal distribution and $\alpha_i={c_{s_i}}/{\left(c_{s_i}+c_{o_i}\right)}$. The optimal-solution function $\theta_*$ is infinitely differentiable with respect to the covariates $x=\left(x_1,\ldots,x_d\right)$. In following experiments, we do not assume that we know the optimal-solution function, and we only use it as the benchmark to compare algorithm performance.

In our experiments, we treat stochastic function inside of the expectation as a simulation black box, where the value and the gradient (with respect to $\theta$) given the decisions $\theta$ and the covariates $x$ may be observable. We apply the OTP framework with all four smoothing techniques to this problem and use the PR-SGD algorithm to solve the offline SO subproblems. Due to space constraints, detailed parameter settings, a description of the PR-SGD algorithm, the implementation of the optimal sample-allocation rules, and the computation of average relative optimality gaps are all provided in Appendix~\ref{app:num}.

We conduct two numerical experiments for this problem. The first evaluates the effectiveness of the optimal sample-allocation rules derived in Section~\ref{sec:analysis}. We consider two instances, $d=2, q=5$ and $d=10, q=5$. In addition to the optimal sample-allocation rule, we examine three fixed-$T$ benchmark rules, where $T$ is set to be $\bar T$, $0.5\bar T$ and $1.5\bar T$. Here, $\bar T=100$ is chosen based on a pilot experiment to ensure that the average relative optimality gap is approximately below $2\%$ (details provided in Appendix{~\ref{Selection_t}). Table~\ref{tab:exp1} reports the average relative optimality gaps for different sample-allocation rules, smoothing techniques, and total budgets $\Gamma$, based on $100$ independent replications, where ``opt'' denotes our optimal sample-allocation rule. Additional results, including standard deviations and ranges of the optimality gaps, are also available in the GitHub repository.

\renewcommand{\arraystretch}{1.1}{
\begin{table}[ht]
\small
    \caption{The Average Relative Optimality Gaps under Different Allocation Rules and Smoothing Techniques}
    \label{tab:exp1}
\begin{tabular}{ccccccccccccc}
\toprule
  Technique  &  & d=2            & \multicolumn{4}{c}{Rule}           &  & d=10             & \multicolumn{4}{c}{Rule}          \\\cmidrule(l){1-1}\cmidrule(l){3-3}\cmidrule(l){4-7}\cmidrule(l){9-9}\cmidrule(l){3-3}\cmidrule(l){10-13}
    &  & $\Gamma$       & opt    & ${\rm T=\bar{T}}$      &  ${\rm T=0.5\bar{T}}$     & ${\rm T=1.5\bar{T}}$    &  & $\Gamma$         & opt   & ${\rm T=\bar{T}}$      & ${\rm T=0.5\bar{T}}$      & ${\rm T=1.5\bar{T}}$    \\\cmidrule(l){1-1}\cmidrule(l){3-3}\cmidrule(l){4-7}\cmidrule(l){9-9}\cmidrule(l){3-3}\cmidrule(l){10-13}
    &  & 250            & 0.3680  & 0.9661  & 0.7787   & 1.6175  &  & $3\times 10^3$   & 0.6587  & 0.8803  & 0.7096  & 0.7937  \\
kNN &  & $2\times 10^3$ & 0.0795  & 0.2211  & 0.1400   & 0.2051  &  & $1.5\times 10^4$ & 0.3778  & 0.3837  & 0.3821  & 0.4603  \\
    &  & $4\times 10^3$ & 0.0507  & 0.0903  & 0.0848   & 0.1292  &  & $3\times 10^4$   & 0.2827  & 0.3199  & 0.2839  & 0.3529  \\\midrule
    &  & 250            & 0.4128  & 0.8820  & 1.1015   & 0.9440  &  & $3\times 10^3$   & 0.7159  & 0.8360  & 0.8101  & 0.7742  \\
KS  &  & $2\times 10^3$ & 0.1038  & 0.1675  & 0.1162   & 0.2311  &  & $1.5\times 10^4$ & 0.4282  & 0.4127  & 0.4250  & 0.4765  \\
    &  & $4\times 10^3$ & 0.0576  & 0.0929  & 0.0654   & 0.0960  &  & $3\times 10^4$   & 0.2560  & 0.3584  & 0.2759  & 0.3725 \\\midrule
    &  & 250            & 0.1502   & -      & - & -      &  & $3\times 10^3$   & 0.0535  & 0.0441  & 0.0535  & -      \\
LR  &  & $2\times 10^3$ & 0.0121  & 0.0154  & 0.0344   & 0.0140  &  & $1.5\times 10^4$ & 0.0024 & 0.0089  & 0.0485  & 0.0037 \\
    &  & $4\times 10^3$ & 0.0064  & 0.0142  & 0.0353   & 0.0093  &  & $3\times 10^4$   & 0.0011  & 0.0083  & 0.0490   & 0.0031 \\\midrule
    &  & 250            & 0.1101 & 0.0506   & 0.0637    & 0.0299    &  & $3\times 10^3$   & 0.0189  & 0.0116  & 0.0495   & 0.0108 \\
KRR &  & $2\times 10^3$ & 0.0049   & 0.0078   & 0.0217   & 0.0063   &  & $1.5\times 10^4$ & 0.0015 & 0.0081  & 0.0469  & 0.0032  \\
    &  & $4\times 10^3$ & 0.0023  & 0.0065  & 0.0207  & 0.0037  &  & $3\times 10^4$   & 0.0009  & 0.0076  & 0.0471  & 0.0028  \\\bottomrule
\end{tabular}
\end{table}}
\vspace{-5pt}

There are several findings from this experiment. (1) The optimal sample-allocation rule outperforms the three fixed-$T$ rules, especially when the budget is sufficiently large. This is consistent with our asymptotic results. Among the fixed-$T$ rules, the choice $T=0.5\bar T$ tends to work well for kNN and KS, whereas $T=1.5\bar T$ performs better for LR and KRR.  (2) kNN and KS yield acceptable results (around a 5\% relative gap) when $d=2$ and the budget is sufficiently large. However, both techniques suffer from the curse of dimensionality and perform poorly when $d=10$. (3) Owing to the infinite differentiability of the optimal-solution function $\theta_*\left( x \right)$, LR and KRR perform very well for this problem, and their performance is not affected by the dimension of the covariates, which agrees with our asymptotic theory. Between the two, KRR outperforms LR because it better captures the nonlinearity of the optimal-solution function.

In the second experiment, we examine the empirical convergence rates of the four smoothing techniques under their respective optimal sample-allocation rules, and study how the dimensions of the decision variable $q$ and the covariate $d$ affect these rates. For each $(d,q)$ setting, we fix a randomly generated online covariate for all four techniques and run $100$ independent replications of the algorithm to estimate the average relative optimality gap, ensuring consistency with the asymptotic results developed in Section~\ref{sec:analysis}. Figure~\ref{fig:convergence} plots the resulting gaps against the total budget $\Gamma$ on a log-log scale for all smoothing techniques and $(d,q)$ combinations.

\vspace{-5pt}
\begin{figure}[th]
    \centering
    \caption{The Average Relative Optimality Gaps under Different $(q,d)$ Settings and Smoothing Techniques}
    \vspace{11pt}
    \label{fig:convergence}
    \includegraphics[width=0.95\linewidth]{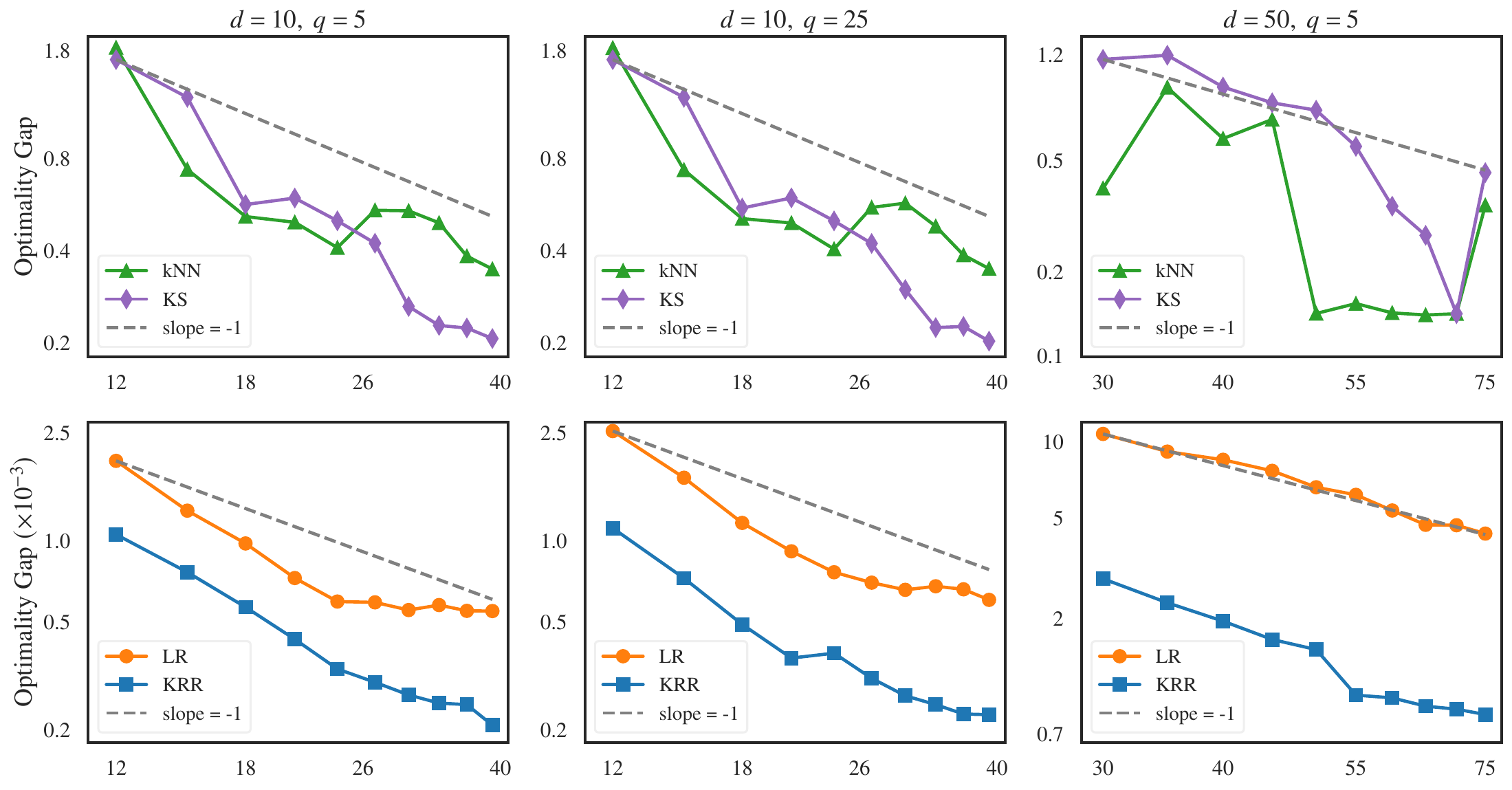}
\end{figure}
\vspace{-20pt}

There are several findings from this experiment. (1) Across all $(d,q)$ settings, LR and KRR produce significantly smaller optimality gaps (note that their gaps are plotted on a $10^{-3}$ scale) with KRR consistently outperforming LR, in line with the first experiment. (2) Comparing the first and second columns of the figure ($d=10$ with $q=5$ and $q=25$), the empirical convergence rates are essentially unchanged, indicating the performance is insensitive to the decision-variable dimension $q$. (3) Comparing the first and third columns ($q=5$ with $d=10$ and $d=50$), kNN and KS show considerable variability and no clear convergence trend, whereas LR and KRR maintain an approximate $\Gamma^{-1}$ rate and remain unaffected by the covariate dimension $d$, making them well suited for high-dimensional settings.

\section{Conclusion}\label{sec:conclusion}

In this paper, we consider contextual strongly convex simulation optimization and propose to solve it using the ``optimization-then-predict" framework where each offline optimization problem is solved with only a finite amount of effort, thus having only an inexact solution. We quantify how inexact solutions affect the convergence rate of the algorithm, and address the critical issue of how to balance the trade-off between the number of offline simulation-optimization subproblems and effort allocated to each subproblem. Though we focus on PR-SGD in solving subproblems and four smoothing techniques (kNN, KS, LR and KRR), our analysis framework may be applied to other simulation-optimization algorithms and other smoothing techniques as well. Numerical results show that the algorithm framework with KRR works remarkably well and avoids the curse of dimensionality, demonstrating its potential in solving high-dimensional problems.

%%%%%%%%%%%%%%%%%

%\ACKNOWLEDGMENT{We would like to express our sincere gratitude to [acknowledge individuals, organizations, or institutions] for their invaluable contributions to this research. We are also grateful to [mention any additional acknowledgements, such as technical assistance, data providers, or colleagues] for their support and assistance throughout the course of this work.}

% References here (outcomment the appropriate case)

% CASE 1: BiBTeX used to constantly update the references
%   (while the paper is being written).
%\bibliographystyle{informs2014} % outcomment this and next line in Case 1
%\bibliography{<your bib file(s)>} % if more than one, comma separated

%\bibliographystyle{informs2014} % outcomment this and next line in Case 1
 % if more than one, comma separated

\bibliographystyle{informs2014}
\bibliography{ref}

% CASE 2: BiBTeX used to generate mypaper.bbl (to be further fine tuned)
%\input{mypaper.bbl} % outcomment this line in Case 2

%If you don't use BiBTex, you can manually itemize references as shown below.

%\bibliographystyle{nonumber}
\newpage

\begin{APPENDICES}

\section{The Smoothness of Optimal-Solution Function}\label{app:smoothness}

\subsection{Proof of Proposition \ref{prop:smoothness}}

\begin{proof}{Proof.}
Fix $x' \in \mathcal{X}$. 
We firstly prove the well-definedness of the optimal-solution function.
By the strong convexity of 
$f\!\left(\cdot;\, x'\right)$ in $\theta$, the solution $\theta_*\!\left(x'\right)$ 
is unique. Hence, the mapping 
$\theta_*:\mathcal{X}\to\Theta$ is single-valued.

To prove the smoothness of the optimal-solution function, we consider the mapping
$J\!\left(\theta, x\right) := \nabla_{\theta} f\!\left(\theta; x\right)$.
By the strong convexity and Assumption~\ref{as:interior}, the optimal solution 
$\theta_*\!\left(x'\right)$ lies in the interior of $\Theta$, so the first-order 
condition \(\nabla_{\theta} f\!\left(\theta_*\!\left(x'\right);\, x'\right) = 0\)
is necessary and sufficient for optimality, implying $J\!\left(\theta_*\!\left(x'\right), \, x'\right) = 0$.
Moreover, the strong convexity ensures that
the Jacobian matrix of $J\!\left(\theta,x\right)$ with respect to $\theta$, i.e.,
$\nabla_\theta J\!\left(\theta,x\right) = \nabla^2_{\theta\theta} f\!\left(\theta_*\!\left(x'\right);\, x'\right)$,
is positive definite and hence invertible.
By Assumption~\ref{as:smoothness}, we also have $J \in C^m\!\left(\Theta \times \mathcal{X}\right)$.
Then, by the 
implicit function theorem \citep{krantz2002implicit}, there exists 
a neighborhood $U' \subseteq \mathcal{X}$ of $x'$ and a unique function 
$g_{x'}: U' \to \Theta$ such that
$J\!\left(g_{x'}\!\left(x\right),\, x\right) =\nabla_\theta f\!\left(g_{x'}\!\left(x\right); x\right)  = 0$ for all $x \in U'$,
and $g_{x'} \in C^m\!\left(U'\right)$.

By the strong convexity and interiority  condition, we also have $\nabla_\theta f\!\left(\theta_*\!\left(x\right); x\right)  = 0$ and $\theta_*\!\left(x\right)$ is the unique solution for each $x \in U'$.
Therefore, we must have 
$\theta_*\!\left(x\right) = g_{x'}\!\left(x\right)$ for all $x \in U'$, which implies that $\theta_*\!\left(x\right)$ is locally 
a $C^m$ function in a neighborhood of $x'$ and this conclusion holds for every $x' \in \mathcal{X}$.

Finally, since these local representations agree on overlaps of neighborhoods 
(again by uniqueness of the minimizer), they patch together to form a globally 
defined function $\theta_*:\mathcal{X}\to\Theta$ that belongs to 
$C^m\!\left(\mathcal{X}\right)$. \hfill\Halmos
\end{proof}

\section{The Optimality Gap and the MSE of the Predicted Solution}

\subsection{Proof of Proposition~\ref{prop:opt_gap}}\label{app:prop_opt_gap}

\begin{proof}{Proof.}
By taking a second-order Taylor's expansion to $f\!\left(\hat{\theta}\!\left(x\right);\, x\right)$ at $\theta_*\!\left(x\right)$, we have
\[
\begin{aligned}
\lefteqn{
f\!\left(\hat{\theta}\!\left(x\right);\, x\right)
-
f\!\left(\theta_*\!\left(x\right);\, x\right)
}\\
 &= \nabla_\theta f\!\left(\theta_*\!\left(x\right);\, x\right)^{\!\top}
\left(
  \hat{\theta}\!\left(x\right)
  -
  \theta_*\!\left(x\right)
\right) 
+ {1\over 2}\left(
  \hat{\theta}\!\left(x\right)
  -
  \theta_*\!\left(x\right)
\right)^{\!\top}
\nabla^{2} f\!\left(\tilde{\theta}\!\left(x\right);\, x\right)
\left(
  \hat{\theta}\!\left(x\right)
  -
  \theta_*\!\left(x\right)
\right)\\
 &= {1\over 2}\left(
  \hat{\theta}\!\left(x\right)
  -
  \theta_*\!\left(x\right)
\right)^{\!\top}
\nabla^{2} f\!\left(\tilde{\theta}\!\left(x\right);\, x\right)
\left(
  \hat{\theta}\!\left(x\right)
  -
  \theta_*\!\left(x\right)
\right),
\end{aligned}
\]
where $\tilde{\theta}\left( x \right)$ lies between $\theta_*\!\left(x\right)$ and $\hat{\theta}\!\left(x\right)$ and the last equality holds according to Assumption \ref{as:interior}. 
Then, according to \cite{boyd2004convex},
\[
\lambda_{\min} \left\| \hat{\theta}\!\left(x\right)-\theta_*\!\left(x\right) \right\|^{2}
\;\leq\;
\left( \hat{\theta}\!\left(x\right)-\theta_*\!\left(x\right) \right)^{\!\top}
\nabla^{2} f\!\left( \tilde{\theta}\!\left(x\right) ; x \right)
\left( \hat{\theta}\!\left(x\right)-\theta_*\!\left(x\right) \right)
\;\leq\;
\lambda_{\max} \left\| \hat{\theta}\!\left(x\right)-\theta_*\!\left(x\right) \right\|^{2},
\]
where $\lambda_{min}$ and $\lambda_{max}$ are the smallest and the largest eigenvalues of $\nabla^2 f\!\left( \tilde{\theta}\!\left( x \right) \right)$. That is, 
\[
\frac{1}{2}\lambda_{\min}\,
\mathbb{E}\!\left[
    \left\|
        \hat{\theta}\!\left(x\right)
        - 
        \theta_*\!\left(x\right)
    \right\|^{2}
\right]
\leq 
\mathbb{E}\!\left[
    f\!\left(\hat{\theta}\!\left(x\right);x\right)
\right]
-
f\!\left(\theta_*\!\left(x\right);x\right)
\leq 
\frac{1}{2}\lambda_{\max}\,
\mathbb{E}\!\left[
    \left\|
        \hat{\theta}\!\left(x\right)
        - 
        \theta_*\!\left(x\right)
    \right\|^{2}
\right].
\]
Since $\nabla^2f\!\left(\tilde{\theta}\!\left(x\right);\, x\right)$ is positive definite due to the strong convexity, we have that all of the eigenvalues are positive, i.e., $\lambda_{min}>0$, and $\lambda_{max}=\left\|\nabla^2f\!\left(\tilde{\theta}\!\left(x\right);\, x\right)\right\|_{op}$ \citep{strang2022introduction}, where $\left\|\cdot\right\|_{op}$ is the operator norm. Due to the the smoothness of $f$ and the compactness of $\Theta$, $\lambda_{max}=\left\|\nabla^2f\!\left(\tilde{\theta}\!\left(x\right);\, x\right)\right\|_{op}$ is bounded and thus
\[
\oE\!\left[f\!\left(\hat{\theta}\!\left(x\right);\,x\right)\right]
-
f\!\left(\theta_*\!\left(x\right);\,x\right)
=
O\!\left(
    \oE\!\left[
        \left\|
            \hat\theta\!\left(x\right)
            -
            \theta_*\!\left(x\right)
        \right\|^{2}
    \right]
\right).
\tag*{\Halmos}
\]
\end{proof}

\subsection{Proof of Corollary~\ref{col:convergence}}\label{app:col_convergence}

\begin{proof}{Proof.}
For any $\eta>0$, by letting $M_f={\oE\left[f\left(\hat\theta\left(x\right);x\right)\right]-f\left(\theta\left(x\right);x\right)\over \eta\oE\left[\left\|\hat\theta\left(x\right)-\theta_*\!\left(x\right)\right\|^2\right]}$, we have
\[
\operatorname{Pr}\left\{{f\left(\hat{\theta}\!\left(x\right);x\right)- f\left(\theta\!\left(x\right);x\right)\over \oE\left[\left\|\hat\theta\!\left(x\right)-\theta_*\!\left(x\right)\right\|^2\right] }\geq M_f\right\}\leq {\oE\left[f\left(\hat{\theta}\!\left(x\right);x\right)\right]- f\left(\theta\!\left(x\right);x\right) \over \oE\left[\left\|\hat\theta\!\left(x\right)-\theta_*\!\left(x\right)\right\|^2\right]M_f}=\eta,
\]
by Markov's inequality. Therefore, $f\left(\hat{\theta}\!\left(x\right);x\right)-f\left(\theta\!\left(x\right);x\right)=O_{\rm P}\left(\oE\left[\left\|\hat\theta\!\left(x\right)-\theta_*\!\left(x\right)\right\|^2\right]\right)$ and $f\left(\hat{\theta}\!\left(x\right);x\right)\to f\left(\theta\!\left(x\right);x\right)$ in probability when $n,T\to\infty$, if $\oE\left[\left\|\hat\theta\!\left(x\right)-\theta_*\!\left(x\right)\right\|^2\right]\to 0$. 
\hfill \Halmos
\end{proof}

\subsection{Proof of Proposition~\ref{prop:MSE_decom}}\label{app:prop_MSE_decom}

\begin{proof}{Proof.}
By the definition of $L_2$ norm, $\oE\left[\left\|\hat\theta\!\left(x\right)-\theta_*\!\left(x\right)\right\|^2\right]$ can be decomposed into $\oE\left[\left\|\hat\theta\!\left(x\right)-\theta_*\!\left(x\right)\right\|^2\right]
=\oE\left[\sum_{j=1}^q\left(\hat\theta^{j}\left( x \right)-\theta_*^{j}\left( x \right)\right)^2\right]
=\sum_{j=1}^q\oE\left[\left(\hat\theta^j\left( x \right)-\theta_*^{j}\left( x \right)\right)^2\right].$

Based on the representation~\eqref{eq:representation}, the bias and variance of $\hat\theta^j\left( x \right),\ j=1,\ldots,q,$ can be written as follows:
\begin{eqnarray*}
\oE\left[\hat{\theta}^j\left( x \right)\right]-\theta^j_*\left( x \right)
&=& \oE\left[\sum_{i=1}^nw\!\left(x_i,\, x\right)\bar{\theta}_T^j\left(x_i\right)\right]-\theta^j_*\left( x \right)
=\sum_{i=1}^nw\!\left(x_i,\, x\right)\left(\oE\left[\bar{\theta}_T^j\left(x_i\right)\right]-\theta_*^j\left(x_i\right)\right)+\sum_{i=1}^nw\!\left(x_i,\, x\right)\theta_*^j\left(x_i\right)-\theta^j_*\left( x \right),\\
\oV\left[\hat{\theta}^j\left( x \right)\right]
&=& \oV\left[\sum_{i=1}^nw\!\left(x_i,\, x\right)\bar{\theta}_T^j\left(x_i\right)\right]
= \sum_{i=1}^nw^2\left(x_i, x\right)\oV\left[\bar{\theta}_T^j\left(x_i\right)\right],
\end{eqnarray*}
where the last equality holds due to the independence of $\bar{\theta}(x_i), \ i=1,\ldots,n$. Then, according to $\oE\left[\left(\hat\theta^j\left( x \right)-\theta_*^{j}\left( x \right)\right)^2\right]=\left(\oE\left[\hat{\theta}^j\left( x \right)\right]-\theta^j_*\left( x \right)\right)^2 + \oV\left[\hat{\theta}^j\left( x \right)\right]$, the stated result can be obtained.
\hfill \Halmos
\end{proof}

\section{The bias order of PR-SGD solution}\label{app:bias}

For simplicity, we omit the covariate $x$ throughout this section, as the SO algorithm is conducted on each fixed $x$.
Our proof strategy of the bias order is motivated by the analytical framework developed in \cite{polyak1992acceleration} for establishing the asymptotic normality of the PR-SGD solution. Instead of analyzing SO probelm directly, \cite{polyak1992acceleration} begin with a general nonlinear root-finding problem of the form $R(\theta)=0$, where $R(\cdot)$ is a nonlinear mapping with root $\theta_*$. The iterative scheme they consider is
\begin{equation}\label{eq:polyak_iter}
\begin{aligned}
&\theta_0\in \Theta\\
&\theta_t = \theta_{t-1}-\gamma_ty_t, \quad y_t = R(\theta_{t-1})+\xi_t, \quad \text{for } t=1,\ldots,T-1,\\
&\bar{\theta}_T = {1\over T}\sum_{t = 0}^{T-1}\theta_t,
\end{aligned}
\end{equation}
with the noise term $\oE\left[\xi_t|\theta_{t-1}\right]=0$. The SO problem $\min_{\theta\in\Theta}f\left(\theta\right)$ naturally fits into this nonlinear root-finding framework: the optimality condition $\nabla f(\theta)=0$ can be viewed as a root-finding problem with $R(\theta_{t-1}) = \oE\left[\Pi_\Theta \left[\nabla_\theta F(\theta_{t-1})\right]\right]$ and $\xi_t = \Pi_\Theta \left[\nabla_\theta F(\theta_{t-1})\right] - \oE\left[\Pi_\Theta \left[\nabla_\theta F(\theta_{t-1})\right]\right]$.
Since the SO problem is a special case of the nonlinear system, Thus, we follow the nonlinear-system perspective of \cite{polyak1992acceleration}, adapting their framework as needed to derive the bias order of the PR-SGD solution. 
To enable a clear and transparent comparison with their original analysis, we retain most of their notation throughout our proof.

\subsection{Regularity conditions}\label{app:bias_conditions}

Our analysis of the bias order builds upon the same regularity conditions employed by \cite{polyak1992acceleration} in their construction of asymptotic normality. Accordingly, we first introduce these conditions, which correspond to Assumptions 3.1–3.4 in \cite{polyak1992acceleration}.

\begin{assumption}\label{as:3.1}
There exists a function $V\left(\theta\right): \Theta \rightarrow \mathbb{R}$ such that for some $\lambda>0$, $\alpha>0, \varepsilon>0, L>0$, and all $\theta_1, \theta_2 \in \Theta$, the conditions $V\left(\theta\right) \geq \alpha\|\theta\|^2,\ \|\nabla V\left(\theta_1\right)-\nabla V(\theta_2)\| \leq$ $L\|\theta_1-\theta_2\|, \ V\left(\theta_*\right)=0, \ \nabla V\left(\theta-\theta_*\right)^T R\left(\theta\right)>0$ for $\theta \neq \theta_*$ hold true. Moreover, $\nabla V\!\left(\theta - \theta_*\right)^{\!\top}
R\!\left(\theta\right)
\;\ge\;
\lambda\, V\!\left(\theta\right)$ for all $\left\|\theta-\theta_*\right\| \leq \varepsilon$.  
\end{assumption}

These regularity conditions are standard in the analysis of optimization.
In the SO problem, we specify $V(\theta) = f(\theta)-f(\theta_*)$, so the assumptions on $V(\cdot)$ reduce to corresponding conditions on the objective function $f(\cdot)$. Under the strong convexity and smoothness assumption, as is illustrated in Assumption~\ref{as:smoothness}, Assumption~\ref{as:3.1} is readily satisfied.

\begin{assumption}\label{as:3.2}
There exists a matrix $G \in R^{N \times N}$ and $K_1<\infty, \delta>0$ 
such that
$$
\left\|R\left(\theta\right)-G\left(\theta-\theta_*\right)\right\| \leq K_1\left\|\theta-\theta_*\right\|^2
$$
for all $\left\|\theta-\theta_*\right\| \leq \delta$ and $\operatorname{Re} \lambda_i\left(G\right)>0, i=1,\ldots,N$.
\end{assumption}

Here, we specify the original condition in \cite{polyak1992acceleration}, which states ``$\left\|R\left(\theta\right)-G\left(\theta-\theta_*\right)\right\| \leq K_1\left\|\theta-\theta_*\right\|^{1+\tau}$, for some $\tau\in(0,1]$'' by setting $\tau=1$.
This choice allows us to obtain an explicit and sharper order for the bias term.
Intrinsically, Assumption~\ref{as:3.2}
controls how well $R(\theta)$ is approximated by its linearization near $\theta_*$, essentially requiring local smoothness of $R(\cdot)$. Under the SO framework, Assumption~\ref{as:3.2} is readily satisfied in the view of Assumption~\ref{as:smoothness}.

\begin{assumption}\label{as:3.3}
$\left(\xi_t\right)_{t \geq 1}$ is a martingale-difference process, i.e., $\oE\left[\xi_t \middle| \mathcal{F}_{t-1}\right]=0$ almost surely, and for some $K_2>0$
$$
\oE\left[\left\|\xi_t\right\|^2 \middle| \mathcal{F}_{t-1}\right]+\left\|R\left(\theta_{t-1}\right)\right\|^2 \leq K_2\left(1+\left\|\theta_{t-1}\right\|^2\right) \quad \text { a.s., }
$$
for all $t \geq 1$. The following decomposition takes place:
$\xi_t=\xi_t\left(0\right)+\zeta_t\left(\theta_{t-1}\right)$,
where
\[
\begin{aligned}
&\oE\left[\xi_t\left(0\right) \middle| \mathcal{F}_{t-1}\right]=0  \quad \text {a.s., } \quad \oE\left[\xi_t\left(0\right) \xi_t^\top\left(0\right) \middle| \mathcal{F}_{t-1}\right] \rightarrow S \quad \text { as } t \rightarrow \infty, \\
&\sup _t \oE\left[\left\|\xi_t\left(0\right)\right\|^2 I\left(\left\|\xi_t\left(0\right)\right\|>C\right) \middle| \mathcal{F}_{t-1}\right] \rightarrow 0 \quad \text { as } C \rightarrow \infty,    
\end{aligned}
\]
for some $S>0$, and, for all $t$ large enough,
$\oE\left(\left\|\xi_t\left(\theta_{t-1}\right)\right\|^2 \middle| \mathcal{F}_{t-1}\right) \leq \delta\left(\theta_{t-1}\right) \ \text {a.s.}$, 
with $\delta\left(\theta\right) \rightarrow 0$ as $\theta \rightarrow 0$.
\end{assumption}

This assumption is about the simulation noise $\xi_t$, which is often thought to be independent random variable with zero mean and finite variance. Therefore, the assumption above can be fulfilled in a general case.

\begin{assumption}\label{as:3.4}
It holds that the stepsize satisfies $\left(\gamma_t-\gamma_{t+1}\right) / \gamma_t=o\left(\gamma_t\right), \gamma_t>0$ for all $t$.
% ;
% $$
% \sum_{t=1}^{\infty} \gamma_t^{{(1+\lambda)/2}} t^{-1 / 2}<\infty
% $$
\end{assumption}

This condition requires the stepsize sequence $\{\gamma_t\}$ to decrease sufficiently slowly. For instance, the commonly used choice $\gamma_t = \gamma t^{-1}$ in basic SGD does not satisfy this restriction. We emphasize that the choice of stepsize plays a critical role in establishing the sharpened bias bound in our analysis.

\subsection{Proof of Proposition \ref{prop:bias}}\label{app:bias_proof}

The proof proceeds as follows. We first construct a linear approximation of the nonlinear iteration~\eqref{eq:polyak_iter} and thus decompose the iteration into its linear component and the corresponding remainder term, following the approach of \cite{polyak1992acceleration}. 
The bias from these two parts are then analyzed respectively. 
Combined with a careful choice of the stepsize order, this yields the overall bias order for the nonlinear system.

\begin{proof}{Proof.}
Define $\Delta_t = \theta_t-\theta_*$ and thus the error of PR-SGD solution  can be written as $\bar{\Delta}_T = \bar{\theta}_T - \theta_* = {1\over T}\sum_{t=0}^{T-1}\left(\theta_t - \theta_*\right) = {1\over T}\sum_{t=0}^{T-1}\Delta_t$.
Thus, deriving the order of bias, $\oE\left[\bar{\theta}_T^j\right]-\theta_*^j$, is equivalent to deriving the order of $\oE\left[\bar{\Delta}_T^j\right]$ for $j=1,\ldots,q$.

Following \cite{polyak1992acceleration}, define $\bar{R}\left(\Delta_t \right) = R\left(\theta_t\right)-R\left(\theta_*\right)=R\left(\theta_t\right)$, where the last equlity holds due to $R(\theta_*)=0$. 
Using this notation, the nonlinear iteration \eqref{eq:polyak_iter} can be rewritten as:
\begin{equation}\label{eq:nonlinear}
\Delta_t 
= \Delta_{t-1}-\gamma_t\left(R\left( \theta_{t-1} \right)+\xi_t\right) 
= \Delta_{t-1}-\gamma_t\left(\bar{R}(\Delta_{t-1})+\xi_t\right).    
\end{equation}

To analyze the bias, we firstly introduce a local linear approximation of $R(\theta)$ by a matrix $G\in \mR^{q\times q}$, as guaranteed by Assumption~\ref{as:3.2}.
Replacing $R(\theta)$ by its its linear approximation $G\left(\theta-\theta_*\right)$ yields the following linear iteration:
\begin{equation*}
\theta_t = \theta_{t-1}-\gamma_t y_t, \quad y_t = G \theta_{t-1}-G\theta_*+\xi_t.
\end{equation*}
Define the corresponding linearized error process by
$\Delta_t^1 = \theta_t-\theta_*$, $\bar{\Delta}_T^1 = {1\over T}\sum_{t=0}^{T-1}\Delta_t^1$ and  $\Delta_0^1 = \Delta_0$.
Then, the linear iteration can be written as
\begin{equation}\label{eq:linear}
\Delta_t^1 
= \Delta_{t-1}^1-\gamma_t\left( G \theta_{t-1} - G\theta_* + \xi_t \right)
= \Delta_{t-1}^1-\gamma_t\left( G \Delta_{t-1}^1 + \xi_t \right).    
\end{equation}

By Lemma 2 of \cite{polyak1992acceleration}, the averaged error of the linear system satisfies
\begin{equation}\label{eq:bar-linear}
\bar{\Delta}_T^1  = {1\over T\gamma_0}\alpha_0 \Delta_0^1 + {1\over T}\sum_{t=0}^{T-1}\alpha_t\xi_t, 
\end{equation}
where $\alpha_t\in\mathbb{R}^{q\times q}$ satisfies $\left\|\alpha_t\right\|\leq K$
for some finite constant $K>0$, and are given by $\alpha_t = \gamma_t\sum_{i=t}^{T-1}\prod_{k=t+1}^i\left(I-\gamma_kG\right)$, with $\prod_{k=t+1}^t\left(I-\gamma_kG\right)=I$.
Note that the boundedness of $\{\alpha_t\}$ follows from Lemma 1 of \cite{polyak1992acceleration} and critically relies on  Assumption~\ref{as:3.4}, which imposes the step-size conditions ensuring stability.

Taking expectations on both sides of Equation~\eqref{eq:bar-linear}, and using the martingale-difference property of $\xi_t$, we obtain the bias order of the linearized system:
\(
\oE\left[\left(\bar{\Delta}_{T}^1\right)^j\right] = O\left(1/T\right),
\)
for $j=1,\ldots,q$.

In the next step, we analyze the bias introduced by the remainder term of the linear approximation.
Define the approximation error
$\delta_T=\bar{\Delta}_T^1- \bar{\Delta}_T$ so that $\oE\left[\bar{\Delta}_T\right] =\oE\left[\bar{\Delta}_T^1\right]- \oE\left[\delta_T\right]$.
Therefore, to establish the order of $\oE\left[\bar{\Delta}_T^j\right]$, it suffices to derive the order of $\oE\left[\delta_T^j\right]$.

According to the proof of Theorem 2 in \cite{polyak1992acceleration}, the approximation error $\delta_T$ satisfies
\[
\begin{aligned}
\left\|\delta_T\right\| \leq {1\over T}\sum_{t=0}^{T-2} K\left\|\bar{R}\left(\Delta_t\right)-G\Delta_t\right\|= \underbrace{{1\over T}\left(K\left\|\bar{R}\left( \Delta_0 \right)-G\Delta_0\right\|+K\left\|\bar{R}\left(\Delta_1 \right)-G\Delta_1\right\|\right)}_{(i)}+
\underbrace{{1\over T}\sum_{t=2}^{T-2} K\left\|\bar{R}\left(\Delta_t\right)-G\Delta_t\right\|}_{(ii)},
\end{aligned}
\]
for the same constant $K>0$ that bounds $\left\{\alpha_t\right\}$.
It's obvious that by the compactness of $\Theta$,
the expectation of term (i) (for each dimension $j=1,\ldots,q$) is of order $O\left(1/T\right)$.

We now focus on term (ii), denote it by $\tilde{\delta}_T$, and proceed to analyze the order of $\oE\left[\tilde{\delta}_T^j\right]$.
The Part 4 of the proof of Theorem 2 in \cite{polyak1992acceleration} gives, that with probability $1$,
\begin{eqnarray}
&& {T\over h(T)}\left\|\tilde{\delta}_T\right\|
\leq {\bar{K}\over h(T)} \sum_{t=2}^{T-2}\left\|\Delta_t\right\|^2
\leq \bar{K}\sum_{t=2}^{\infty}{\left\|\Delta_t\right\|^2\over h(t)}\label{eq:tilde_delta}\\
&& \sum_{t=2}^{\infty}{\oE\left[\left\|\Delta_t\right\|^2\right]\over h(t)}\leq \tilde{K}\sum_{t=2}^{\infty}{\gamma_t\over h(t)}\label{eq:gamma},
\end{eqnarray}
where $h(t)$ is nondecreasing in $t$ and satisfies $h(T)=o(T)$, and
$\bar{K}, \tilde{K}>0$ are constants.

At this point, the choice of $h(t)$ and the step-size sequence $\{\gamma_t\}$ is crucial for pinning down the bias order.  
By specifying $h(t)=\left(\log t\right)^3$ and $\gamma_t = {\log t\over t}$ (satisfying Assumption~\ref{as:3.4}) and plug them into Equation~\eqref{eq:gamma}, we obtain
\[
\sum_{t=2}^{\infty}{\oE\left[\left\|\Delta_t\right\|^2\right]\over \left( \log t \right)^3}\leq \tilde{K}\sum_{t=2}^{\infty}{1\over \left( \log t \right)^2 t }<+\infty.
\]
Then, by Kronecker's lemma \citep{shiryaev1996probability}, we obtain
\[
\lim_{T\to \infty}{1\over \left(\log T\right)^3}\sum_{t=2}^{T-2}\left( \log t \right)^3 {\oE\left[\|\Delta_t\|^2\right]\over \left( \log t \right)^3}
=\lim_{T\to \infty}{1\over \left(\log T\right)^3}\sum_{t=2}^{T-2} \oE\left[\|\Delta_t\|^2\right]
= 0.
\]
Together with \eqref{eq:tilde_delta}, this yields
\[
{T\over \left(\log T\right)^3} \oE\left[\left\|\tilde{\delta}_T\right\|\right]
\leq {\bar{K}\over \left(\log T\right)^3}\sum_{t=2}^{T-2} \oE\left[\|\Delta_t\|^2\right]\to 0, \text{ as } T\to \infty.
\]
Hence, $\oE\left[\left\|\tilde{\delta}_T\right\|\right] = O\left({\left(\log T\right)^3\over T}\right)$ and thus $\oE\left[\tilde \delta_T^j\right] = {O}\left({\left(\log T\right)^3\over T}\right)= \tilde{O}\left(1/T\right)$ for $j=1,\ldots,q$.

Combining the expectation of term (i) and (ii) yields that the bias introduced by the remainder term is 
\[
\oE\left[\delta_T^j\right] = O\left({1\over T}\right) + \oE\left[\tilde \delta_T^j\right] = \tilde O\left({1\over T}\right),\ j=1,\ldots,q.
\]

Finally, combining the the bias from the linear approximation and the remainder term leads to
$$\oE\left[\bar{\theta}_T^j\right]-\theta_*^j = \oE\left[\bar{\Delta}_T^j\right] =\oE\left[\left(\bar{\Delta}_T^1\right)^j\right]-\oE\left[\delta_T^j\right]=O\left(1\over T\right) + \tilde{O}\left(1\over T\right) = \tilde{O}\left(1\over T\right),\ j =1,\ldots,q,$$  
which completes the proof.\hfill
\Halmos
\end{proof}

\section{The Proofs of the Smoothing Techniques}\label{app:smoothing}

We first state the following lemma, which will play a key role in analyzing the error of $k$NN and the variance of KS.
\begin{lemma}\label{lem:counting}
Let $\mathcal{X}\subset\mathbb{R}^d$ be a compact set
% bounded domain satisfying a volume regularity condition, 
and let $\{x_i\}_{i=1}^n\subset\mathcal{X}$ be a quasi-uniform design with separation distance $q_n$ and fill distance $h_n$.
Suppose that the target point $x\in\mathcal{X}$ is an interior point.
% , and the radius $r\to 0$ as $n\to\infty$.
Denote $N\!\left(x,r\right)=\#\left\{i:\|x_i-x\|\leq r \right\}$ as the number of design points within distance $r$ from $x$. For $r> h_n$.
We have $N\!\left(x,r\right)\asymp n r^d$, i.e., $r\asymp \left(N\!\left(x,r\right)/n\right)^{1/d}$.
\end{lemma}

\begin{proof}{Proof.}
We establish the order of $N\!\left(x,r\right)$ by deriving its upper and lower bounds, respectively.

Let $V_d$ be the volume of the unit ball in $\mathbb{R}^d$.
For the upper bound, note that if $\|x_i-x\|\leq r$, then balls $B\left(x_i, q_n\right)$ is contained in $B\left(x, r+q_n\right)$.
Hence, $\bigcup_{\|x_i-x\|\leq r}B\left(x_i, q_n\right)\subset B\left(x, r+q_n\right)$.
Furthermore, by the definition of separation distance, balls $B\left(x_i, q_n\right)$ are disjoint.
Taking volumes leads to $N\!\left(x,r\right)q_n^dV_d \leq \left( r + q_n \right)^d V_d$, i.e., $N\!\left(x,r\right)\leq \left(1+r/q_n\right) ^d$.

We then derive the lower bound. 
For any $y\in B\left(x, r - h_n\right)$, there exists $x_j$ such that $\|y-x_j\|\leq h_n$, by the definition of fill distance. Then, $\|x-x_j\|\leq \|x-y\|+\|y-x_j\|\leq \left(r - h_n\right)+h_n=r$, i.e., such $x_j$ satisfies $\|x-x_j\|\leq r$.
Thus, every $y\in B\left(x, r - h_n\right)$ can be covered by some ball $B(x_j, h_n)$ with $\|x_j-x\|\leq r$. 
Therefore, $B\left(x, r - h_n\right)\subset \bigcup_{\|x_i-x\|\leq r} B(x_i,h_n)$. Taking volumes yields $\left(r - h_n\right)^d V_d\leq N\!\left(x,r\right)h_n^d V_d$, i.e., $N\!\left(x,r\right)\geq \left( r/h_n - 1 \right) ^d$.

Combining the two bounds, we obtain
$
\left( r/h_n - 1 \right) ^d\leq N\!\left(x,r\right)\leq\left(1+r/q_n\right) ^d.
$
Under quasi-uniform design, i.e., $h_n\asymp q_n\asymp n^{-1/d}$, this implies
$nr^d\lesssim N\!\left(x,r\right)\lesssim nr^d$, which is followed by 
$N\!\left(x,r\right)\asymp nr^d$ or $r\asymp (N\!\left(x,r\right)/n)^{1/d}$.\hfill\Halmos
\end{proof}

\begin{remark}\label{remark}
\textit{By Lemma~\ref{lem:counting}, we can deduce some critical results for kNN and KS.
For kNN, by letting $N\!\left(x,r\right)=k$, we have $r\asymp \left(  k/n \right)^{1/d}$ under the condition that $k\to\infty$ and $k/n\to 0$ as $n\to\infty$, i.e., the $k$ nearest neighbors of $x$ are contained in a ball of radius of order $\left(  k/n \right)^{1/d}$.
For KS with bandwidth $h$, by letting $r=h$, we have $N\left(x,h\right)\asymp nh^d$ under the condition that $h\to 0$ as $n\to\infty$, i.e.,  the number of the designed points falling in the band $\|x_i-x\|\leq h$ is of order $nh^d$.
Lemma~\ref{lem:counting} also reveals a  connection between kNN and KS.
Specifically, KS with bandwidth $h$ can be viewed as an $n h^d$-nearest neighbor technique, while kNN with $k$ neighbors can be regarded as KS with bandwidth of order $\left(  k/n \right)^{1/d}$.
% In this correspondence, the asymptotic conditions $k\to\infty$ and $k/n\to 0$ in $k$NN are equivalent to requiring $n h^d \to \infty$ and $h\to 0$ in KS, respectively.
}
\end{remark}

\subsection{Proof of k-Nearest Neighbors (kNN)}\label{app:kNN}

\subsubsection{The Optimal Allocation and Convergence Rate Under CR setting}

Following the similar steps as the proofs of Lemma~\ref{lem:kNN} and Theorem~\ref{thm:kNN}, except assuming $\bar{\theta}^j_T\left( x \right)$ is unbiased for all $x\in\mathcal{X}$ (which makes the solution-induced bias zero), the bias and the variance of $\theta_{kNN}^j\left( x \right)$ are of order $O\left(\left({k/ n}\right)^{1/d}\right)$ and $O\left({1\over kT}\right)$, respectively. Therefore, the optimal sample-allocation rule is to balance these two terms, leading to $n^*\asymp\Gamma^{d\over d+2}k^*$, $T^*\asymp\Gamma^{2\over d+2}/k^*$, $(k^*)^{-1}=o(1)$ and $k^*\lesssim \Gamma^{2\over d+2}$.
That is, the order of $T^*$ lies in the interval $\left[1,\Gamma^{{2\over d+2}}\right)$ and $n^*$ lies in the interval $\left(\left.\Gamma^{{2\over d+2}},\Gamma\right]\right.$.
 Note that there is no lower bound for the order of $T^*$.
The corresponding optimal convergence rate of the optimality gap is of order $O_{\rm P}\left(q\Gamma^{-\frac{2}{d+2}}\right)$.

\subsection{Proof of Kernel Smoothing (KS)}\label{app:KS}

\subsubsection{Proof of Lemma~\ref{lem:KS}}\label{app:KS_lem}

\begin{proof}{Proof.}
We first analyze the bias of $\hat{\theta}_{KS}\left( x \right)$. Based on Equation~\eqref{eq:KS}, the bias term is
\[
\oE\left[\theta_{KS}^j\left( x \right)\right]-\theta_*^j\left( x \right)
= \oE\left[\sum_{i=1}^n w\!\left(x_i,\, x\right)\bar{\theta}_T^j\left(x_i\right)\right]-\theta_*^j\left( x \right)   
= \sum_{i=1}^n w\!\left(x_i,\, x\right) \left(\oE\left[\bar{\theta}^j_T\left(x_i\right)\right]-\theta^j_*\left(x_i\right)\right)
+\sum_{i=1}^n w\!\left(x_i,\, x\right)\left(\theta^j_*\left(x_i\right)-\theta^j_*\left( x \right)\right) ,
\]
where $w\!\left(x_i,\, x\right)=I\{\|x_i-x\|\leq h\}/\sum_{i=1}^nI\{\|x_i-x\|\leq h\}$.
By Proposition~\ref{prop:bias}, we have that solution-induced bias is
$\sum_{i=1}^n w\!\left(x_i,\, x\right)\left(\theta^j_*\left(x_i\right)-\theta^j_*\left( x \right)\right)=\tilde{O}\left(1/T\right)$ as $\sum_{i=1}^n w\!\left(x_i,\, x\right)=1$.
By $L$-Lipchitz continuous as is illustrated in the proof of Lemma~\ref{lem:kNN}, 
the interpolation bias $\left|\theta_*^j\left(x_i\right)-\theta_*^j\left( x \right)\right|\leq L\|x_i-x\|\leq Lh$ for $x_i$ such that $\|x_i-x\|\leq h$.
Therefore, the overall bias is $\oE\left[\theta_{KS}^j\left( x \right)\right]-\theta_*^j\left( x \right) = \tilde{O}\left(1/T+h\right)$.

Subsequently, we analyze the variance term:
\[
\oV\left[\theta^j_{KS}\left( x \right)\right]
= \oV\left[\sum_{i=1}^n w\!\left(x_i,\, x\right)\bar{\theta}_T^j\left(x_i\right)\right]
= \sum_{i=1}^n w\!\left(x_i,\, x\right)^2\oV\left[\bar{\theta}_T^j\left(x_i\right)\right].
\]
By defining $N\left(x,h\right)=\#\{i: \|x_i-x\|\leq h\}$,  the weight $w\!\left(x_i,\, x\right)=I\{\|x_i-x\|\leq h\}/\sum_{i=1}^nI\{\|x_i-x\|\leq h\}$ can be rewritten as: $w\!\left(x_i,\, x\right)=1/N\left(x,h\right)$ if $\|x_i-x\|\leq h$ while $w\!\left(x_i,\, x\right)=0$ if $\|x_i-x\|>h$.
Thus, 
\[
\sum_{i=1}^n w\!\left(x_i,\, x\right)^2\oV\left[\bar{\theta}_T^j\left(x_i\right)\right]
= \sum_{\|x_i-x\|\leq h} \left({1\over N\left(x,h\right)}\right)^2 \oV\left[\bar{\theta}_T^j\left(x_i\right)\right] 
% = \oV\left[\bar{\theta}_T^j\left(x_i\right)\right]/N\left(x,h\right)
= O\!\left(\frac{1}{T n h^d}\right),
\]
by Lemma~\ref{lem:counting} and Remark~\ref{remark}, which demonstrates $N\left(x,h\right)\asymp nh^d$.

Therefore, $\oE\left[\left(\hat{\theta}_{KS}^j\left( x \right)-\theta^j_*\left( x \right)\right)^2\right] 
= \tilde{O}\left(h^2+{1\over T^2}\right) + O\left({{1\over Tnh^d}}\right)
= \tilde{O}\left(h^2+{1\over T^2}+{1\over Tnh^d}\right), \quad j=1,\ldots,q. $\hfill\Halmos
\end{proof}

\subsubsection{Proof of Theorem \ref{thm:KS}}\label{app:KS_thm}

\begin{proof}{Proof.}
For a fixed computation budget $\Gamma$, the optimal convergence rate of the MSE can be obtained by solving the optimization problem as follows:
\begin{eqnarray*}
\min_{n,T,h}\ h^2+{1\over T^2}+{1\over Tnh^d}, \quad
 \text{subject to}\quad nT=\Gamma,\ h=o(1),\ \left(nh^d\right)^{-1}=o(1).
\end{eqnarray*}
Notice that the squared interpolation bias $h^2$ and the variance term ${1\over Tnh^d} = {1\over \Gamma h^d}$ do not depend on the allocation scheme of $n$ and $T$. 
Balancing these two terms yields $h^* \asymp \Gamma^{-{1\over d+2}}$ and $\left( h^*\right)^2+{1\over \Gamma \left( h^*\right)^d} \asymp \Gamma^{-{2\over d+2}}$.

To ensure that the overall MSE is not dominated by the solution-induced error $\tfrac{1}{T^2}$, we require $T\gtrsim \Gamma^{{1\over d+2}}$.
Meanwhile, the constraint $\left(nh^d\right)^{-1}=o(1)$ requires $n^{-1}=o(h^{d})=o\left(\Gamma^{-{d\over d+2}}\right)$.
Therefore, the order of $T^*$ lies in the interval $\left.\left[\Gamma^{1\over d+2}, \Gamma^{2\over d+2}\right)\right.$ and $n^*$ lies in the interval $\left.\left(\Gamma^{d\over d+2}, \Gamma^{d+1\over d+2}\right]\right.$
with the resulting optimal MSE of order $\tilde{O}\left(\Gamma^{-\tfrac{2}{d+2}}\right)$.
Similarly, by Proposition~\ref{prop:MSE_decom} and Corollary~\ref{col:convergence}, the optimality gap converges in the rate of $\tilde O_{\rm P}\left(q\Gamma^{-\tfrac{2}{d+2}}\right)$.
 \hfill\Halmos
\end{proof}

\subsubsection{The Optimal Allocation and Convergence Rate Under CR setting}

Following the similar steps as the proofs in Appendices~\ref{app:KS_lem} and \ref{app:KS_thm}, except assuming $\bar{\theta}^j_T\left( x \right)$ is unbiased for all $x\in\mathcal{X}$ (which makes the solution-induced bias zero), the bias and the variance of $\theta_{KS}^j\left( x \right)$ are of order $O\left(h\right)$ and $O\left({1\over Tnh^d}\right)$, respectively. Therefore, the optimal sample-allocation rule is to balance these two terms, leading to $h^*\asymp\Gamma^{-{1\over d+2}}$ and $\left({n^*}\right)^{-1}=o(h^{d})=o\left(\Gamma^{-{d\over d+2}}\right)$.
Explicitly, the order of $T^*$ lies in the interval $\left.\left[1,\Gamma^{{2\over d+2}}\right.\right)$ and $n^*$ lies in the interval $\left.\left(\Gamma^{{2\over d+2}},\Gamma\right.\right]$.
Note that there is no lower bound for the order of $T$.
The corresponding optimal convergence rate of the optimality gap is of order $O_{\rm P}\left(q\Gamma^{-\tfrac{2}{d+2}}\right)$.

\subsection{Proof of  Linear Regression (LR)}\label{app:LR}

\subsubsection{Proof of Lemma~\ref{lem:LR}}\label{app:LR_lem}

\begin{proof}{Proof.}
We first analyze the bias of $\theta_{LR}^j\left( x \right)$. Based on Equation~\eqref{eq:LR}, the bias term is 
\[
\oE\left[\theta_{LR}^j\left( x \right)\right]-\theta_*^j\left( x \right)
=w\left( x \right)^\top\oE\left[\bar{\theta}_T^j\right]-\theta_*^j\left( x \right)
=w\left( x \right)^\top\left(\oE\left[\bar{\theta}_T^j\right]-\theta_*^j\right)
+\left(w\left( x \right)^\top \theta_*^j-\theta_*^j\left( x \right)\right),
\]
where $w\left( x \right) = \Phi\left(\Phi^\top\Phi\right)^{-1}\phi\left( x \right)\in \mathbb{R}^n$,$\bar{\theta}_T^j = (\bar{\theta}^j_T\left(x_1\right), \ldots, \bar{\theta}^j_T\left( x_n \right))^\top$ and $\theta^j_* = (\theta^j_*\left(x_1\right), \ldots, \theta^j_*\left( x_n \right))^\top$. 
Let $\mu^\top=\frac{1}{n}\bm 1_n^\top \Phi\in\mathbb{R}^s$ and $ 
A=\frac{1}{n}\Phi^\top\Phi\in\mathbb{R}^{s\times s}.$
Then,
$
\bm 1_n^\top w\left( x \right)=\bm 1_n^\top \Phi(\Phi^\top\Phi)^{-1}\phi\left( x \right)
= (n\mu^\top)\,(nA)^{-1}\phi\left( x \right)=\mu^\top A^{-1}\phi\left( x \right).
$
Assuming the basis is bounded and $A
$ is well-conditioned (i.e., $\|A^{-1}\|_{\rm op}=O(1)$), we obtain
$
|\bm 1_n^\top w\left( x \right)|\;\le\;\|\mu\|\,\|A^{-1}\|_{\rm op}\,\|\phi\left( x \right)\| \;=\; O(1).
$
Thus, according to Proposition~\ref{prop:bias}, the solution-induced bias
\[
w\left( x \right)^\top\left(\oE\left[\bar{\theta}_T^j\right]-\theta_*^j\right)
=\tilde{O}\left({1\over T}\right)w\left( x \right)^\top \bm 1_n = \tilde{O}\left({1\over T}\right).
\]

The interpolation bias is 
\[
w\left( x \right)^\top \theta_*^j-\theta_*^j\left( x \right)
= \left(\theta_*^j\right)^\top w\left( x \right)-\theta_*^j\left( x \right)
=\left(\Phi \beta^j\right)^\top \Phi\left(\Phi^\top\Phi\right)^{-1}\phi\left( x \right)
-\left(\beta^j\right)^\top\phi\left( x \right)
=0
\]
by plugging $\theta_*^j=\Phi \beta^j$ and $\theta_*^j\left( x \right)=\left(\beta^j\right)^\top\phi\left( x \right)$.
The overall bias is $\oE\left[\theta_{LR}^j\left( x \right)\right]-\theta_*^j\left( x \right)=\tilde{O}\left(1/T\right)$.

Subsequently, we analyze the variance term:
\[
\oV\left[\theta_{LR}^j\left( x \right)\right]
= \oV\left[w\left( x \right)^\top\bar{\theta}_T^j\right]
=w\left( x \right)^\top \oV\left[\bar{\theta}_T^j\right]w\left( x \right)
\asymp {1\over T}w\left( x \right)^\top I w\left( x \right).
\]
Notice that  $w\left( x \right)^\top I w\left( x \right) = \phi\left( x \right)^\top \left(\Phi^\top\Phi\right)^{-1}\phi\left( x \right)=O(1/n)$ as ${1\over n}\Phi^\top \Phi = {1\over n}\sum_{i=1} \phi\left(x_i\right)\phi\left(x_i\right)^\top\to\Sigma$ for some $\Sigma$, the variance
$\oV\left[\theta_{LR}^j\left( x \right)\right]\asymp {1\over nT}$.

Therefore, $\oE\left[\left(\hat{\theta}_{LR}^j\left( x \right)-\theta^j_*\left( x \right)\right)^2\right] 
= \tilde{O}\left({1\over T^2}\right) + O\left({{1\over nT}}\right)
= \tilde{O}\left({1\over T^2}+{1\over nT}\right), \quad j=1,\ldots,q. $\hfill\Halmos
\end{proof}

\vspace{10pt}

\begin{remark}
\textit{If the empirical basis functions are orthonormalized, i.e., 
$\langle\phi_l,\phi_j\rangle_n = \tfrac{1}{n}\sum_{i=1}^n \phi_l\left(x_i\right)\phi_j\left(x_i\right)=1$ for $l=j$ and 
$\langle\phi_l,\phi_j\rangle_n=0$ for $l\neq j$, $l,j=1,\ldots,s$, then 
${1\over n}\Phi^\top\Phi = I_s$ and the weight vector is 
$w\left( x \right) = \tfrac{1}{n}\Phi \phi\left( x \right)$, with the $i$-th entry given by 
$w\!\left(x_i,\, x\right)={1\over n}\langle\phi\left( x \right),\phi\left(x_i\right)\rangle$.  Under this representation, the solution-induced bias and variance simplify as follows:  
\[
w\left( x \right)^\top\!\left(\mathbb{E}\left[\bar{\theta}_T^j\right]-\theta_*^j\right)
= \frac{1}{n} \sum_{i=1}^n \langle \phi\left( x \right),\phi\left(x_i\right)\rangle 
\left(\mathbb{E}\left[\bar{\theta}_T^j\left(x_i\right)\right]-\theta_*^j\left(x_i\right)\right)
= \tilde{O}\!\left(\tfrac{1}{T}\right),
\]
where the last equality holds because each term in parentheses is $\tilde{O}\left(1/T\right)$ and the weights are uniformly bounded.   Similarly, the variance of the LR estimator satisfies
\[
\oV\!\left[\hat\theta_{LR}^j\left( x \right)\right]
= \sum_{i=1}^n w\!\left(x_i,\, x\right)^2\,\oV\!\left[\bar{\theta}_T^j\left(x_i\right)\right]
= \frac{1}{n^2}\sum_{i=1}^n \langle\phi\left( x \right),\phi\left(x_i\right)\rangle^2 \, O\!\left(\tfrac{1}{T}\right)
= O\!\left(\tfrac{1}{nT}\right).
\]}
\end{remark}

\begin{remark}
\textit{Furthermore, if the basis function is augmented with an intercept term, i.e., 
\(\phi\left( x \right) = (\phi_1\left( x \right), \dots, \phi_s\left( x \right))^\top \in \mathbb{R}^s\) with \(\phi_1\left( x \right)\equiv 1\), 
then we can show that the weights sum to one: 
\(\bm 1_n^\top w\left( x \right)=\sum_{i=1}^n w\left(x, x_i\right)=1\), similar to the case of kNN and KS.  }
\textit{Specifically, let $e_1 = (1,0,\ldots,0)\in\mathbb{R}^s$ and thus $\Phi^\top \bm 1_n = \Phi^\top\Phi e_1$. Then
\[
\bm 1_n^\top w\left( x \right) 
= \bm 1_n^\top \Phi\left(\Phi^\top\Phi\right)^{-1}\phi\left( x \right)
= e_1 \Phi^\top\Phi \left(\Phi^\top\Phi\right)^{-1}\phi\left( x \right) =1.
\]
With the normalization $\bm 1_n^\top w\left( x \right)=1$, the solution-induced bias simplifies to
\[
w\left( x \right)^\top\!\left(\mathbb{E}\!\left[\bar{\theta}_T^j\right]-\theta_*^j\right)
= \sum_{i=1}^n w\!\left(x_i,\, x\right)\left(\mathbb{E}\left[\bar{\theta}_T^j\left(x_i\right)\right]-\theta_*^j\left(x_i\right)\right),
\]
which has the same order as the individual solution bias terms, i.e., 
\(\asymp \mathbb{E}\left[\bar{\theta}_T^j\left(x_i\right)\right]-\theta_*^j\left(x_i\right)\) for $i=1,\ldots,n$.  }
\end{remark}

\subsubsection{Proof of Theorem \ref{thm:LR}}\label{app:LR_thm}
\begin{proof}{Proof.}
For a fixed computation budget $\Gamma$, the optimal convergence rate of the MSE can be obtained by solving the optimization problem as follows:
\begin{eqnarray*}
\min_{n,T}\ {1\over T^2}+{1\over Tn}, \quad
 \text{subject to}\quad nT=\Gamma.
\end{eqnarray*}
To ensure that the overall MSE is not dominated by the solution-induced bias, it only requires that $T^2\gtrsim \Gamma$, i.e., $T^*\gtrsim\Gamma^{1/2}$.
The corresponding optimal order of MSE is $\tilde{O}\left({1/\Gamma}\right)$.
Similarly, by Proposition~\ref{prop:MSE_decom} and Corollary~\ref{col:convergence}, the optimality gap converges in the rate of $\tilde O_{\rm P}\left(q\Gamma^{-1}\right)$.

\end{proof}

\subsubsection{The Optimal Allocation and Convergence Rate Under CR setting}
Following the similar steps as above, except $\bar{\theta}^j_T\left( x \right)$ is unbiased for all $x\in\mathcal{X}$, there is no bias and the variance is of order $O\left({1\over \Gamma}\right)$. Therefore, there is no bias-variance trade-off and the allocation between $n$ and $T$ is free. The corresponding optimal convergence rate of the optimality gap is of order $O_{\rm P}\!\left(q\Gamma^{-1}\right)$.

\subsection{Proof of Kernel Ridge Regression (KRR)}\label{app:KRR}

\subsubsection{The Optimal Allocation and Convergence Rate Under CR setting}
Following the similar steps as proofs of Lemma~\ref{lem:KRR} and Theorem~\ref{thm:KRR}, except $\bar{\theta}_T^j\left( x \right)$ is unbiased for all $x\in\mathcal{X}$, the bias and the variance of $\hat\theta_{KRR}\left( x \right)$ are of order $O\left(\lambda^{1-{d\over 2m}}\right)$ and $O\left(\Gamma^{-1}\lambda^{-{d\over2m}}\right)$, respectively, and the MSE is of order $O\left(\lambda^{1-{d\over 2m}}+\Gamma^{-1}\lambda^{-{d\over2m}}\right)$.
The optimal choice is $\lambda^*\asymp \Gamma^{-1}$ and $T^*=o\left(\Gamma^{1-{d\over2m}}\right)$, and the corresponding optimal convergence rate of the optimality gap is of order $O_{\rm P}\left(q\Gamma^{-1+{d\over 2m}}\right)$. 

\vspace{11pt}

\section{Supplementary Details of Numerical Study}\label{app:num}

\subsection{Detailed Parameter Settings}

We impose homogeneous weights $w_{ij}=1$ and identical shortage and overage cost parameters $c_{s_i}=c_s=3$ and $c_{o_i}=c_o=1$ for all items $i$. To ensure positive demand, we set  $\gamma=0.3$. The domain of the covariate vector is $X \in [0,3]^d$. The idiosyncratic mean demands $\mu_i$ lie in $[0,0.4]$ and are equally spaced, i.e., $\mu_i \in \{\,0,\, 0.4/(q-1),\,\ldots,\,0.4\,\}$.

\newpage
For both experiments, the test covariates are generated uniformly from the interior of the domain, that is, $X_{\mathrm{test}} \sim \mathrm{Unif}([0.1,2.9]^d)$. This interiority maintains consistency with the theoretical requirements in Lemma~\ref{lem:counting}, and allows the performance of the nonparametric smoothing techniques without boundary effects.

\subsection{Implementation of PR-SGD}\label{app:prsgd}

\textbf{Stochastic-gradient derivation.}
Denote the stochastic function inside the expectation of the objective as
\[
F\left(\theta; x\right)=\sum_{i=1}^q \left(c_{s}(D_i\left( x \right)-\theta_i)^+ + c_{o}(\theta_i-D_i\left( x \right))^+\right).
\]
For each component $i$, direct differentiation yields
\begin{equation}\label{eq:newsvendor_SGD}
{\partial F\over\partial \theta_i}= -c_s \mathbf{1}\{D_i\left( x \right)>\theta_i\} + c_o \mathbf{1}\{D_i\left( x \right)\leq \theta_i\},  
\end{equation}
and thus the stochastic gradient is 
$\nabla_\theta F\left(\theta; x\right)=\left({\partial F\over\partial \theta_1},\ldots, {\partial F\over\partial \theta_q}\right)$.

Since $D_i = \sum_{j=1}^d Z_j + \varepsilon_i,\ i=1,\ldots,q,$ conditional on a fixed covariate $x$, we draw fresh samples of $Z_1,\ldots,Z_d$ (covariate-driven common factors) and $\varepsilon_1,\ldots,\varepsilon_q$ (idiosyncratic noise terms) at each PR-SGD iteration to obtain a new realization of $D$, and then compute the stochastic gradient using Equation~\eqref{eq:newsvendor_SGD}.

\textbf{Step-size choice.}
In our experiments, we adopt the step-size of order $O\left(\log t/t\right)$ to remain consistent with the bias analysis in Proposition~\ref{prop:bias}; see Appendix~\ref{app:bias_proof} for details.

\textbf{Algorithmic implementation of PR-SGD.}
For a fixed covariate $x$, the implementation of PR-SGD Algorithm in this newsvendor example is summarized below.

\begin{algorithm}[H]
\small
\caption{PR-SGD Algorithm for a fixed covariate $x$ in Newsvendor Example}
\begin{algorithmic}[1]
\State \textbf{Input:} total iterations $T$, feasible region $\Theta$ (with $\theta\ge 0$), step-size sequence $\{\gamma_t\}$.
\State \textbf{Initialize:} choose any $\theta_0\left( x \right)\in\Theta$.
\For{$t=1$ to $T$}
    \State Generate observation of $Z_1,\ldots,Z_d$ and $\varepsilon_1,\ldots,\varepsilon_q$ and compute the demand observation $D$ conditional on $x$.
    \State Compute the stochastic gradient
$\nabla_\theta F\left(\theta_{t-1}\left( x \right);x\right)$ by Equation~\eqref{eq:newsvendor_SGD}.
\State Update $\theta_t\left( x \right) = \Pi_\Theta\!\left[\theta_{t-1}\left( x \right) - \gamma_t \nabla_\theta F\left(\theta_{t-1}\left( x \right);x\right)\right]$ with $\gamma_t = \gamma \frac{\log(t+1)}{t+1}$, where $\gamma>0$ is a constant.
\EndFor
\State \textbf{Output:} averaged solution $\bar{\theta}_T\left( x \right) = \frac{1}{T+1}\sum_{t=0}^{T} \theta_t\left( x \right).
$
\end{algorithmic}
\end{algorithm}

\subsection{Implementation of Optimal Sample-Allocation Rules}

The optimal sample-allocation rules in our experiments follow the theoretical prescriptions in Theorems~\ref{thm:kNN}--\ref{thm:KRR}. 
The choices of $(n,T)$ depend on the smoothing technique, the dimension $d$, and the smoothness $m$ (see Table~\ref{table:allocation}). 
For the newsvendor experiments with $d\in\{2,10,50\}$, the allocations are specified as follows.

\textbf{(1) kNN.}
The optimal allocation satisfies $T^*\in\left.\left[\Gamma^{1/\left(d+2\right)},\,\Gamma^{2/\left(d+2\right)}\right.\right)$, so the choice of $T$ depends strongly on the dimension $d$.  In particular, the required solution exactness decreases in high-dimensional settings, i.e.,  $T$ is chosen smaller when $d$ is large. 
For $d=2$, since $T^*\in\left.\left[\Gamma^{1/4},\,\Gamma^{1/2}\right.\right)$, we let $T$ and $n$ be of comparable magnitude so that $T^*\approx\sqrt{\Gamma}$.  
For $d=10$, where $T^*\in\left.\left[\Gamma^{1/12},\,\Gamma^{1/6}\right.\right)$, we let $T$ grow slowly.  
For $d=50$, where $T^*\in\left.\left[\Gamma^{1/52},\,\Gamma^{1/26}\right.\right)$, we keep $T$ essentially constant given the finite range of $\Gamma$.

\textbf{(2) KS.}
The optimal allocation $T^*\in\left.\left[\Gamma^{1/\left(d+2\right)},\,\Gamma^{2/\left(d+2\right)}\right.\right)$ is same as kNN, we use the same allocation rules as kNN.

\textbf{(3) LR.}
The optimal allocation satisfies $T^*\in\left.\left[\Gamma^{1/2},\,\Gamma\right.\right]$ and thus solution exactness is beneficial. 
We keep $n$ fixed (but larger than the number of basis functions) and let $T$ scale on the same order as $\Gamma$ for all dimensions.

\textbf{(4) KRR.}
The optimal allocation satisfies $T^*\in\left.\left[\Gamma^{1/2-d/\left(4m\right)},\,\Gamma^{1-d/\left( 2m\right)}\right.\right)$. 
In our newsvendor example, the optimal solution is infinitely smooth, so we may take $m\to\infty$, under which the dependence on $d$ disappears. 
Accordingly, we choose $T^*$ close to order $\Gamma$ (but slightly below), while letting $n$ increase only very slowly, across all dimensional settings.

\subsection{The Selection of $\bar{T}$ in the Fixed-$T$ Benchmark Rules}\label{Selection_t}

A practical implementation of the fixed-$T$ benchmark rules for OTP is to compute sufficiently accurate solutions at the $n$ design points using a constant number of simulation iterations $\bar{T}$. 
To select an appropriate value of $\bar{T}$, we conduct a pilot study as follows.

We randomly sample $1024$ covariate vectors from the covariate domain $\mathcal{X}$ and run PR-SGD with a given number of iteration steps.  
For each covariate, we compute its relative optimality gap, and then take the average of these $1024$ gaps.  
This gives one estimate of the average error corresponding to the chosen $\bar{T}$.  

We repeat this entire procedure with $100$ independent replications, and use the grand average (i.e., the mean of the $100$ averaged gaps) as the performance metric.  
We then examine whether this grand average is below the $2\%$ threshold.

For $d=2$, using $\gamma=0.8$ and $100$ iteration steps, the grand average relative optimality gap is approximately $1.74\%$.  
For $d=10$, using $\gamma=4$ and $100$ iteration steps, the grand average relative optimality gap is approximately $1.93\%$.  
Based on these pilot results, we set $\bar{T}=100$ in our numerical experiments.

\subsection{Computation of Relative Optimality Gap Statistics}\label{Optimality_Statistics}
The results reported in Table~\ref{tab:exp1} are based on $100$ online test covariate vectors to ensure robustness. 
Specifically, we uniformly draw $100$ covariates from the domain $\mathcal{X}$. 
Given the budget $\Gamma$, the smoothing technique, and the sample-allocation rule, we execute OTP at each covariate and compute its relative optimality gap.
We then summarize these $100$ gaps by their mean, standard deviation, minimum, and maximum, thereby assessing both the average performance and the stability of OTP.

To further reduce statistical variability, we repeat the entire procedure above for $100$ independent replications.  
Each replication yields four summary statistics (mean, standard deviation, minimum, and maximum) based on the $100$ covariates.  
We then compute the grand average of these statistics across the $100$ replications.

Table~\ref{tab:exp1} reports the grand average of the mean relative optimality gap for selected budget levels $\Gamma$.  
Due to page limitations, the corresponding results for the standard deviation, minimum, and maximum gaps, as well as the $(n,T)$ allocations for additional budget values, are provided in the accompanying GitHub repository at \url{https://github.com/NiffyLin/CSCSO}.

In fact, the additional summary statistics -- standard deviation, minimum, and maximum -- lead to the same conclusions as Table~\ref{tab:exp1}.  
They further demonstrate the advantage of the derived optimal allocation rules.  
Across all budget levels, the optimal allocation rule typically delivers the most stable performance, as reflected by its smaller standard deviation, minimum and maximum of relative optimality gaps.  
Among the four smoothing methods, KRR achieves the most accurate and stable results, and LR also performs well, whereas kNN and KS deteriorate substantially in high-dimensional settings.

\end{APPENDICES}

\end{document}